\def\year{2021}\relax
\documentclass[letterpaper]{article} 
\usepackage{aaai21}  
\usepackage{times}  
\usepackage{helvet} 
\usepackage{courier}  
\usepackage[hyphens]{url}  
\usepackage{graphicx} 
\usepackage{natbib}  
\usepackage{caption} 
\usepackage[ruled,vlined]{algorithm2e}
\usepackage{soul}

\usepackage{graphicx}
\usepackage{colortbl}
\usepackage{multirow}
\usepackage{amsmath}
\usepackage[switch]{lineno} %
\title{How Robust are Model Rankings : A Leaderboard Customization Approach for Equitable Evaluation}

\author {
        Swaroop Mishra\textsuperscript{\rm 1}, Anjana Arunkumar\textsuperscript{\rm 1}
}

\affiliations{

    \textsuperscript{\rm 1}School of Computing, Informatics, and Decision Systems Engineering\\
    Arizona State University, Tempe, AZ 85281, USA\\
    \texttt\{ srmishr1, aarunku5\}@asu.edu

}

\begin{document}

\maketitle

\begin{abstract}
Models that top leaderboards often perform unsatisfactorily when deployed in real world applications; this has necessitated rigorous and expensive pre-deployment model testing. A hitherto unexplored facet of model performance is: \textit{Are our leaderboards doing equitable evaluation?} In this paper, we introduce a task-agnostic method to probe leaderboards by weighting samples based on their `difficulty' level. We find that leaderboards can be adversarially attacked and top performing models may not always be the best models. We subsequently propose alternate evaluation metrics. Our experiments on 10 models show changes in model ranking and an overall reduction in previously reported performance-- thus rectifying the overestimation of AI systems' capabilities. Inspired by behavioral testing principles, we further develop a prototype of a visual analytics tool that enables leaderboard revamping through customization, based on an end user's focus area. This helps users analyze models' strengths and weaknesses, and guides them in the selection of a model best suited for their application scenario. In a user study, members of various commercial product development teams, covering 5 focus areas, find that our prototype reduces pre-deployment development and testing effort by 41 \% on average.
\end{abstract}

\section*{Introduction}
Machine Learning (ML) models have achieved super-human performance on several popular benchmarks such as Imagenet \cite{russakovsky2015imagenet}, SNLI \cite{bowman2015large} and SQUAD \cite{rajpurkar2016squad}. Models selected based on their success in leaderboards of existing benchmarks however, often fail when deployed in real life applications. A hitherto unexplored question that might answer why our `optimal' model selection is sometimes found to be `sub-optimal' is: \textit{Are our leaderboards doing equitable evaluation?} There are two aspects associated with model evaluation-- performance scores and ranking. Existing evaluation metrics have been shown to inflate model performance, and overestimate the capabilities of AI systems \cite{recht2019imagenet,sakaguchi2019winogrande, bras2020adversarial}. Ranking on the other hand, has remained unexamined.

Consider that Model \textit{X} is ranked higher than Model \textit{Y} in a leaderboard. \textit{X} could be solving all `easy' questions and fail on all `hard' questions, while \textit{Y} solves some `hard' questions but a lesser number of questions overall. Should \textit{Y} be ranked higher than \textit{X} then? The ranking depends on the application scenario-- does the user want a model to compulsorily answer the hardest questions or answer more `easy' questions? 

Question `difficulty' demarcation is application dependent. For example, in sentiment analysis for movie reviews, if a model learns to associate `not' with negative sentiment, a review containing the phrase `not bad' might be incorrectly labeled. Such spurious bias-- unintended correlation between model input and output \cite{torralba2011unbiased, bras2020adversarial}-- makes samples `easier' for models to solve. Similarly, in conversational AI, models are often trained on clear questions (`where is the nearest restaurant'), but user queries are seen to often diverge from this training set \cite{kamath2020selective} (`I'm hungry'). Such atypical queries have higher out-of-distribution (OOD) characteristics, and are consequently `harder' for models to solve. 

\textit{How do we quantify and acknowledge `difficulty' in model evaluation?} Recently, semantic-textual-similarity (STS) with respect to the training set has been identified as an indicator of OOD characteristics \cite{Mishra2020OurEM}. Similarly, we propose 2 algorithms to quantify spurious bias in a sample. We also propose \textit{WSBias} (Weighted Spurious Bias), an equitable evaluation metric that assigns higher weights to `harder' samples, extending the design of \textit{WOOD} (Weighted OOD) Score \cite{Mishra2020OurEM}. 

A model's perception of sample `difficulty' however, is based on the confidence it associates with its answer for each question. Models' tendency to silently answer incorrectly, with high confidence, hinders their deployment. For example, if Model \textit{X} incorrectly answers questions with high confidence, but has higher accuracy than Model \textit{Y} which has low confidence for incorrect answers, then Model \textit{Y} is preferable, especially in safety critical domains. We quantify a model's view of difficulty using maximum softmax probability-- a strong baseline which has been used as a threshold in selective answering \cite{hendrycks2016baseline}-- to construct \textit{WMProb} (Weighted Maximum Probability), a metric that assigns higher weights to samples solved with higher confidence.

Based on these considerations, we make the following novel contributions to enable better model selection for deployment in real life applications:

\begin{figure}[ht]
    \centering
    \includegraphics[width=0.47\textwidth]{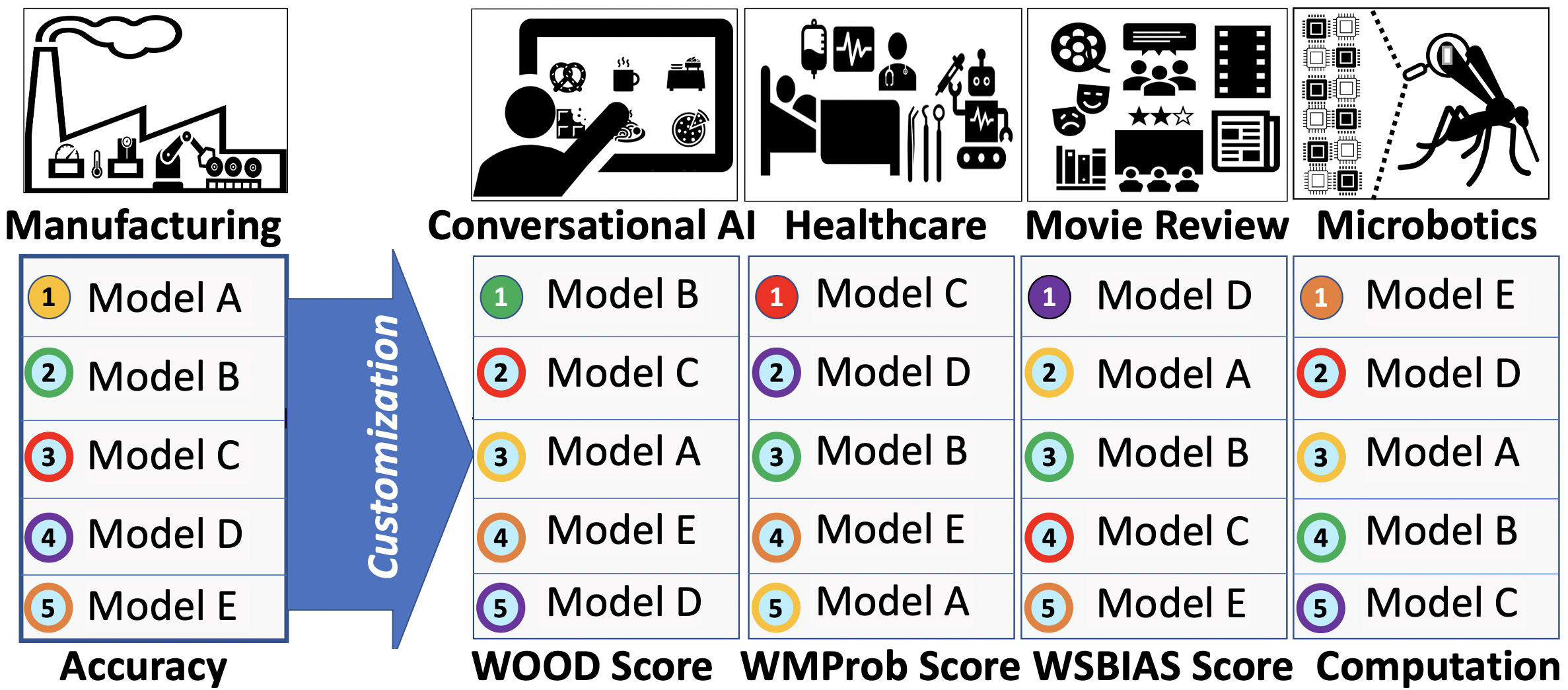}
    \caption{Leveraging various evaluation metrics for leaderboard customization, depending on application requirements. A robot in a manufacturing industry \cite{heyer2010human}-- which is a less risky environment, compared to say healthcare-- is unlikely to experience OOD data, computational constraints, and spurious bias. Thus accuracy can be used for evaluation.}
    \label{introfig}
\end{figure}

\begin{itemize}
    \item We present a methodology to probe leaderboards and check if `better' models are ranked higher.
    \item  We propose alternate evaluation metrics for leaderboard customization based on application specific requirements, as shown in Figure \ref{introfig}. To the best of our knowledge, we are the first to propose leaderboard customization.
    \item Our automatic sample annotation approach based on `difficulty' indicators speeds up conventional evaluation processes in all focus areas. For example, current OOD generalization evaluation has the overhead of identifying OOD datasets corresponding to each dataset (without a consistent guideline distinguishing IID from OOD); OOD datasets might also have their own bias. Automatic annotation with STS allows for the identification of sample subsets that exhibit greater OOD characteristics, thus enabling the use of in-house data splits for evaluation.
    \item In order to help users select models best suited to their applications we leverage ideas from software engineering \cite{beizer1995black,Ribeiro2020BeyondAB} and develop a tool prototype that interactively visualizes how model rankings change, based on flexible weight assignment to different splits of data. 
    \item We perform a user study with experts from industries with different focus areas, and show that our prototype helps in calibrating the proposed metrics by reducing development and testing effort on average by 41\%.
    \item We show how the metrics reduce model performance inflation, minimizing the overestimation of capability of AI.
    \item Our analysis yields some preliminary observations of the strengths and weaknesses of 10 models; we recommend the use of appropriate evaluation metrics to do fair model evaluation, thus minimizing the gap between research and the real world. 
\end{itemize}

\section*{Experimental Setup}
We experiment with ten different models-- Bag-of-Words Sum (BOW-SUM) \cite{harris1954distributional}, Word2Vec Sum (W2V SUM) \cite{mikolov2013distributed}, GloVe Sum (GLOVE SUM) \cite{pennington2014glove}, Word2Vec LSTM (W2V LSTM) \cite{hochreiter1997long}, GloVe LSTM (GLOVE LSTM), Word2Vec CNN (W2V CNN) \cite{lecun1995convolutional}, GloVe CNN (GLOVE CNN), BERT Base (BERT BASE) \cite{devlin2018bert}, BERT Large (BERT LARGE) with GELU \cite{hendrycks2016gaussian}, and RoBERTa Large (ROBERTA LARGE) \cite{liu2019roberta}, following a recent work on OOD Robustness \cite{hendrycks2020pretrained}. We analyze these models over two movie review datasets: (i) SST-2 \cite{socher2013recursive}-- contains short movie reviews written  by experts , and (ii) IMDb \cite{maas2011learning}-- has full-length lay movie reviews. We train models on SST-2 and evaluate on both SST-2 and IMDb. This setup ensures both IID (SST-2 test set) and OOD (IMDb) evaluation. 

\section*{Metric Formulation}

Based on our three-pronged quantification of difficulty, we adapt \textit{WOOD} Score, and propose \textit{WSBias} and \textit{WMProb}.

\subsection{Ranking Samples:} 

\noindent\textbf{\textit{WSBias}:} Samples are ranked in decreasing order of the amount of spurious bias present. Algorithms \ref{algo:one} and \ref{algo:two} elaborate our approach to quantify the presence of spurious bias across samples. Algorithm \ref{algo:one} is inspired from Curriculum Learning \cite{xu2020curriculum} and AFLite \cite{sakaguchi2019winogrande, bras2020adversarial}-- a recent technique for adversarial filtering of dataset biases. Algorithm \ref{algo:two} incorporates bias related to train and test set overlap \cite{lewis2020question, gorman2019we}. Spurious bias here represents the ease (\% of times) with which a sample is correctly predicted by linear models on top of RoBERTA features, irrespective of inter-model agreement used in measuring sample uncertainty \cite{lung2013assessing, chen2014projection}.

\noindent\textbf{\textit{WOOD}:} First, STS is calculated for each test set value with respect to all train set samples. The STS values for the varying percentages of the train set data-- ranging from the top 1\%-100\% of the train set, in turn obtained by sorting the train set samples in descending order of STS against each test set sample-- with respect to each test set sample are averaged. The test samples are then sorted in descending order based on this averaged STS value. This is done as train-test similarity is a task dependent hyper-parameter, which can lead to either inductive or spurious bias \cite{Mishra2020DQIMD, gorman2019we}. 

\noindent\textbf{\textit{WMProb}:} Test samples are ranked in increasing order, based on the confidence of model prediction for that sample (i.e., the maximum softmax probability). WMProb differs from the previous two metrics in that it operates on a model-dependent parameter, i.e., prediction  confidence, rather than data-dependent (and model-independent) parameters. 

\begin{algorithm}[ht]

\SetAlgoLined
\KwResult{Input: Testset $T$, Hyper-Parameters $m$, $t$, Models $M$-[Logistic Regression, SVM], Output: Spurious bias values $B$ for each sample $S$}
Fine-tune RoBERTA on 10\% of $T$ and discard this 10\% to get $R$ and find $R$'s embeddings $e$\;
Evaluation Score $E(S)=0$ for all $S$ in $R$ \;
Correct Evaluation Score $C(S)=0$ for all $S$ in $R$ \;
 \ForAll{$i \in m$}{
 Randomly select trainset of size $t$ from $R$ \;
 \While {$y<$ 2}
 {
 Train $M$[$y$] on $t$ using $e$ and evaluate on rest of $R$ i.e. $V$ \;
 \ForAll{$S \in V$}{ $E(S) = E(S)+1$\; \If {model prediction is correct}
 {$C(S) = C(S)+1$}}
 }
}
 \ForAll{$S \in T$}{$B(S)=C(S)/E(S)$}
 \caption{Bias within Test Set}
 \label{algo:one}
\end{algorithm}
\begin{algorithm}[ht]
\SetAlgoLined
\KwResult{Input: Trainset $Tr$, Testset $T$, $Z$=\{Logistic Regression, SVM Linear, SVM RBF, Naive Bayes\} and Output: Spurious bias values $B$ for each sample $S$}
Correct Evaluation Score $C(S)=0$ for all $S$ in $T$ \;
 \ForAll{$i \in Z$}{
 Train Model $Z$ on $Tr$ and Evaluate on $T$ \;
 \ForAll{$S \in T$}{ \If {model prediction is correct}
 {$C(S) = C(S)+1$}}
 }
 \ForAll{$S \in T$}{$B(S)=C(S)/4$}
 \caption{Bias across Train and Test Set}
 \label{algo:two}
\end{algorithm}

\subsection{Split Formation and Weight Assignment:} The dataset is divided into splits, based either on user defined thresholds of the sample `difficulty', or such that the splits are equally populated (Figure \ref{fig:pcp}). Weight assignment for samples can either be done continuously, or split-wise (Table \ref{tab:1}). We also assign penalties to incorrectly answered samples by multiplying the assigned sample weights with penalty scores-- as defined below. 

\subsection{Formalization:}
\label{ssec:form}
Let $D$ represent a dataset where $T$ is the test set and $Tr$ is the train set. $B$ is the degree of bias / averaged STS value that a sample has, $p$ is the model prediction, $g$ is the gold annotation, $a$ ($=1$ in our experiments) is a hyper-parameter used in continuous weight calculation, $b_1$ and $b_2$ are the weights assigned per split, $th_1$ is the splitting parameter, either based on threshold value or equally populated splits. $d$ and $e$ are the reward and penalty scores respectively, which are each multiplied with both $b_{1}$ and $b_{2}$ across splits. Here, $W_{continuous}$ and $W_{two-split}$ refer to continuous and two-split weighting schemes respectively.

\begin{equation}
W_{continuous}=\frac{a}{B}
\label{eq:1}
\end{equation}

\begin{equation}
W_{two-split}=
    \begin{cases}
      b_1, & \text{if}\ B > th_1  \text{  (Split 1)}\\
      b_2, & \text{otherwise   (Split 2)}
    \end{cases}
\label{eq:2}
\end{equation}
\begin{equation}
    K=
    \begin{cases}
      d, & \text{if}\ p=g \\
      e, & \text{otherwise}
    \end{cases}
\end{equation}
\begin{equation}
Metric=\frac{\sum_{T}K*W}{\sum_{T}d*W}*100
\end{equation}

When $B$ is the maximum softmax prediction probability, the value of $W_{continuous}$ must be reciprocated in equation \ref{eq:1}, and the split assignment must be swapped in equation \ref{eq:2}.

\subsection{Reward and Penalty Scores:}
For each sample, $d$ and $e$ are respectively used as reward and penalty multipliers of the assigned weights. This is done in order to flexibly score each sample's contribution to the overall performance. For example, in Table \ref{tab:1}, $b_1=1$ for Split 1, $b_2=2$ for Split 2. In Case 1, $d=1$ and $e=-1$, meaning that in Split 1, correct and incorrect samples are assigned scores of $b_1*d=1$ and $b_1*e=-1$ respectively; in Split 2, this follows as $b_2*d=2$ and $b_2*e=-2$. $d$ and $e$ can vary, as shown across the other cases. We also use quantified spurious bias, average STS, and maximum softmax probability values for $d$ and $e$, as shown in Cases 6-9.

\section*{Leaderboard Probing Results}
We use \textit{WSBias}, \textit{WOOD}, and \textit{WMProb} Scores to probe accuracy-based model performance/ranking. Based on the metric selected, we gain insights of models' behavior over the two datasets considered, for different aspects of sample `difficulty', in terms of overall/split-wise performance.

\begin{table*}[t]
\centering
\begin{tabular}{|c|c|c|c|c|c|c|c|c|c|}
\hline
\multirow{2}{*}{\textbf{Case}} & \multirow{2}{*}{\textbf{Description}} & \multirow{2}{*}{\textbf{$b_1$}} & \multirow{2}{*}{\textbf{$b_2$}} & \multirow{2}{*}{\textbf{d}} & \multirow{2}{*}{\textbf{e}} & \multicolumn{2}{c|}{\textbf{Split 1}} & \multicolumn{2}{c|}{\textbf{Split 2}} \\ \cline{7-10} 
 &  &  &  &  &  & \textbf{Correct} & \textbf{Incorrect} & \textbf{Correct} & \textbf{Incorrect} \\ \hline
1 & Reward = Penalty & 1 & 2 & 1 & -1 & 1 & -1 & 2 & -2 \\ \hline
2 & Reward Only & 1 & 2 & 1 & 0 & 1 & 0 & 2 & 0 \\ \hline
3 & Penalty Only & 1 & 2 & 0 & -1 & 0 & -1 & 0 & -2 \\ \hline
4 & Reward $>$ Penalty & 1 & 2 & 1 & -0.5 & 1 & -0.5 & 2 & -1 \\ \hline
5 & Penalty $>$ Reward & 1 & 2 & 0.5 & -1 & 0.5 & -1 & 1 & -2 \\ \hline
6 & Continuous Weights & $1/B$ & $1/B$ & 1 & -1 & $1/B$ & $-1/B$ & $1/B$ & $-1/B$ \\ \hline
7 & Continuous Weights (*) & $B$ & $B$ & 1 & -1 & $B$ & $-B$ & $B$ & $-B$ \\ \hline
8 & Reward = Penalty = $B$ & 1 & 2 & $1/B$ & $-1/B$ & $1/B$ & $-1/B$ & $2/B$ & $-2/B$ \\ \hline
9 & Reward = Penalty = $B$ (*) & 1 & 2 & $B$ & $-B$ & $B$ & $-B$ & $2B$ & $-2B$ \\ \hline
\end{tabular}
    \caption{Weighting Schemes Tested for \textit{WSBias}, \textit{WOOD}, and \textit{WMProb}. Here, examples of weighting schemes for 2 splits are shown, with Split 1 containing the `easy' samples, and Split 2 containing the `hard' samples. In Cases 6 and 8, $1/B$ refers to quantified sample `difficulty' and applies to \textit{WSBias} and \textit{WOOD} Score. In Cases 7 and 9, $B$ is the `difficulty' for \textit{WMProb}.}
        \label{tab:1}

\end{table*}

\subsection{Range of Hyper-Parameters}
We vary split numbers from 2 (as in Equation \ref{eq:2}) to 7, which are either equally populated, or formed based on equally spaced thresholds. We use the splits in combination with all weighting schemes (as in Table \ref{tab:1}) to calculate the metric values. Higher split numbers have additional weight and threshold hyper-parameters. For example, in the case of 4 splits, we have weight hyper-parameters of $b_{1}-b_{4}$ and threshold hyper-parameters of $th_{1}$-$th_{3}$. We multiply each weight by $d$ or $e$ to assign sample scores. For example, the respective correct/incorrect sample scores per split when $b_{1}-b_{4}$ take values from $1-4$, and $d=1$, $e=-0.5$ are: +1/-0.5,+2/-1,+3/-1.5,+4/-2. We also vary the $b_{n}$ hyper-parameter values along linear (additive and subtractive), logarithmic, and square scales. Results for SST-2 show lesser extents of performance inflation and ranking change-- though these are still significant-- which we attribute to SST-2 being the IID dataset. We report results for 7 splits over the IMDb Dataset below; we note that similar results are observed over different hyper-parameter considerations for both datasets.

\subsection{WSBias}

\textbf{GLOVE-LSTM Outperforms ROBERTA-LARGE in `Easy' Questions for Certain Splits:}
We see that when data is divided into 2-5 splits, ROBERTA-LARGE, despite having higher accuracy, has greater, equal, and lesser errors in Splits 1 (easiest) - 3 respectively, than GLOVE-LSTM. This indicates that ROBERTA-LARGE is unable to solve easy samples efficiently. 

\noindent\textbf{Transformers Fail to Answer `Easy' Questions:}
In Figure \ref{fig:pcp}, in the parallel coordinates plot (PCP) model performance values vary to a greater extent in Splits 1-3, and converge till they are near identical in Splits 6-7. Transformer models (BERT-BASE, BERT-LARGE, ROBERTA-LARGE) display poor performance  on Split 1, with only ROBERTA-LARGE having a marginally positive value. This may be attributed to catastrophic forgetting, which occurs due to the larger quantities of data transformer models are trained on to improve generalization.

\noindent\textbf{Frequent Ranking Changes Across Splits for Word Averaging and GLOVE Embedding Models:}
In the PCP, the occurrence of crossed/overlapping lines indicates frequent ranking changes for the BOW-SUM, GLOVE-LSTM, and GLOVE-CNN models over all the splits. This result is replicated for all weighting schemes.

\noindent\textbf{Significant Changes in Model Ranking:}
In the multi-line chart (MLC) (Figure \ref{fig:mlc}), the scores do not monotonically increase unlike the behavior seen in accuracy. There is a significant change in model ranking, with 8/10 models changing positions, as indicated by the red dots. The concurrence of significant ranking changes using both algorithms indicates that irrespective of model training, accuracy does not appropriately rank models based on spurious bias considerations.

\noindent\textbf{Significant Reduction in Model Performance Inflation:}
The overall \textit{WSBias} score values decrease by: (i) ~25\%-63\% for Algorithm \ref{algo:one}, and (ii)~11\%-40\% for Algorithm \ref{algo:two}, with respect to accuracy. This indicates that \textit{WSBias} solves accuracy's inflated model performance reporting.

\noindent\textbf{Transformer Model Performance is the Least Inflated:}
We observe in the MLC that the performance drops seen with \textit{WSBias} are the least for transformer models, and the most for word-averaging models. This follows from the insight that transformer models predominantly fail on `easier' samples, i.e., those which contain greater amounts of spurious bias. As `hard' samples-- with  low  spurious  bias--  are  predominantly  negatively scored for simpler models, the overall \textit{WSBias} Score of the model decreases. This is because the absolute weights assigned for those samples are higher. Our findings thus indicate that simpler models are over-reliant on spurious bias.

\subsection{WOOD}

\noindent \textbf{Model Performance Inflation Increases as \% of Training Set Used to Calculate STS Increases:}
We note that overall model performance increases when larger percentages of the training set are used in the calculation of averaged STS values for test samples. This is due to the corresponding decrease in the variance of the averaged STS values (values tend towards 0.5 at 100\%).

\noindent \textbf{Split-wise Ranking Varies More for Lesser \%s of Training Set Used to Calculate STS:}
Due to the decreased variation effect of averaging STS values for higher percentages 

\begin{figure*}[t]
    \centering
    \includegraphics[width=0.8\textwidth]{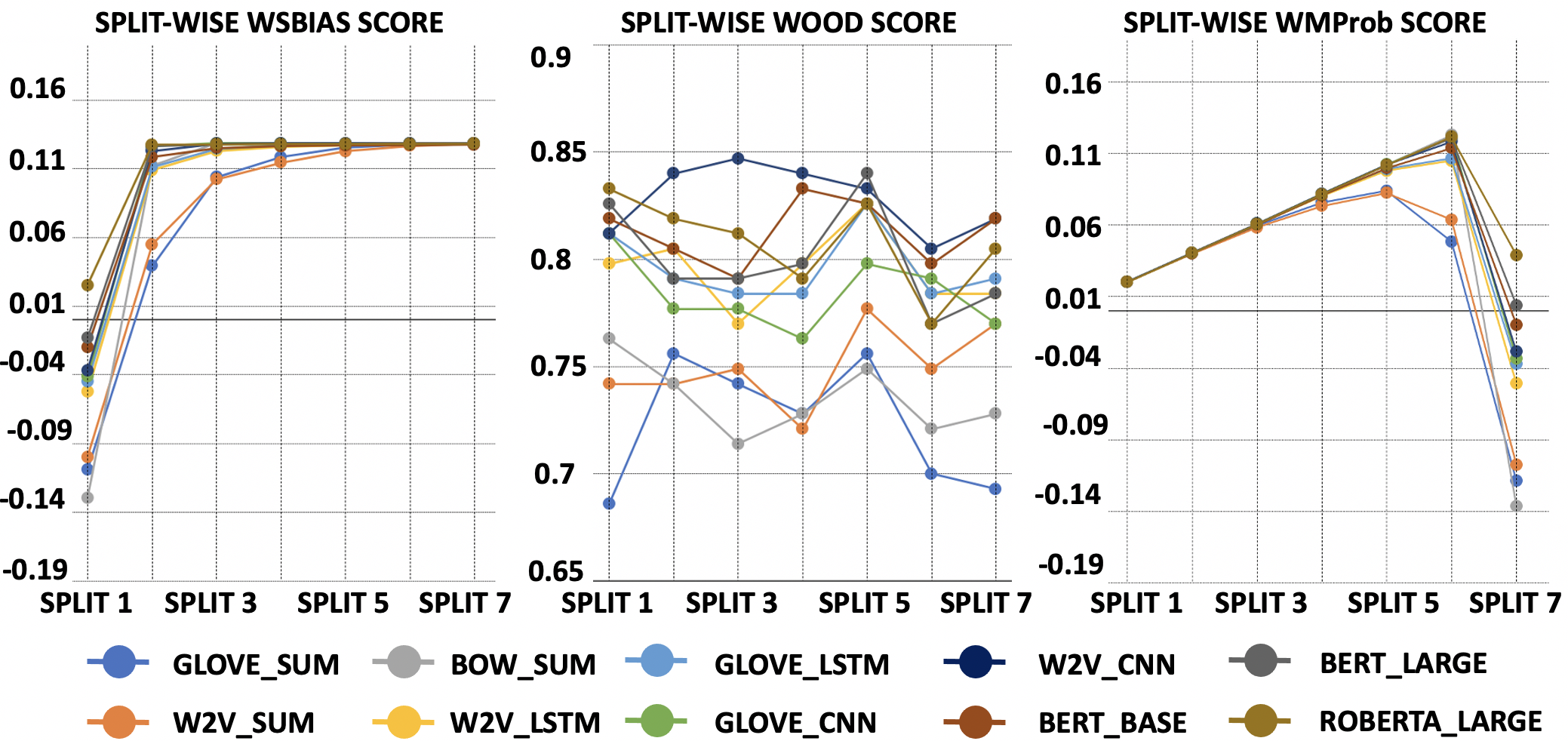}
    \caption{Split-wise results for each model using \textit{WSBias} (based on Algorithm \ref{algo:one} -- left), \textit{WOOD} (based on top 25\% STS averaging -- middle), and \textit{WMProb} (right) are  shown in the PCPs. Each vertical line indicates the data split considered (Split 1 -- Easiest/Lowest Confidence, Split 7 -- Hardest/Highest Confidence). Weighting scheme considered is Case 1, for 7 equally populated splits on the IMDb dataset}
    \label{fig:pcp}
\end{figure*}

\noindent($>50$), there is less fluctuation in the split-wise distributions of incorrect predictions, especially in thresholded splits. As a result, the split-wise model ranking changes more frequently when lower percentages are used for STS calculation. In Figure \ref{fig:pcp} (25\%), this is indicated by clear and frequent line crossings.

\noindent \textbf{Overall Ranking Changes Significantly for All \%s of Training Set Considered:} 
There is significant ranking change irrespective of STS \%, to a greater extent than with \textit{WSBias}. For example, in Figure \ref{fig:mlc}, we see that (i) 9/10 models change their ranking position based on WOOD (calculated with 25\% STS), and (ii) there is a decrease of ~15\% in WOOD score values, with respect to accuracy. These respectively indicate that accuracy does not rank appropriately in the context of model generalization, and also inflates model performance on OOD datasets. These results are replicated over all weight and STS \% considerations.

\noindent \textbf{W2V-CNN Beats Transformers in Most Splits:}
 The transformer models are consistently surpassed by W2V-CNN in Splits 2,3,4,6,7. W2V-CNN also beats transformers in overall \textit{WOOD} Score; this is because transformers fail to solve samples with low STS (i.e., higher OOD characteristics, absolute weight assignments),and are thus heavily penalized. 
 
\subsection{WMProb}

\noindent \textbf{Some Models are Better Posed to Abstain than Others:}
When splits are formed based on both threshold and equal population constraints, the number of incorrect predictions is approximately equally distributed across splits for all 
models. In particular, ROBERTA-LARGE, BERT-LARGE, and GLOVE-SUM display near-equal numbers of incorrect samples at the `easiest' (high confidence) and `hardest' (low confidence) splits. On the other hand, BERT-BASE, GLOVE-LSTM, and W2V-LSTM show near-monotonic decrease in the number of samples correctly answered from high to low confidence. For safety critical applications, models are made to abstain from answering if the maximum softmax probability is below a specified threshold. If a lesser number of incorrect predictions (i.e., higher accuracy) is associated with high model confidence, that model is preferred for deployment. 

\begin{figure*}[t]
    \centering
    \includegraphics[width=0.87\textwidth]{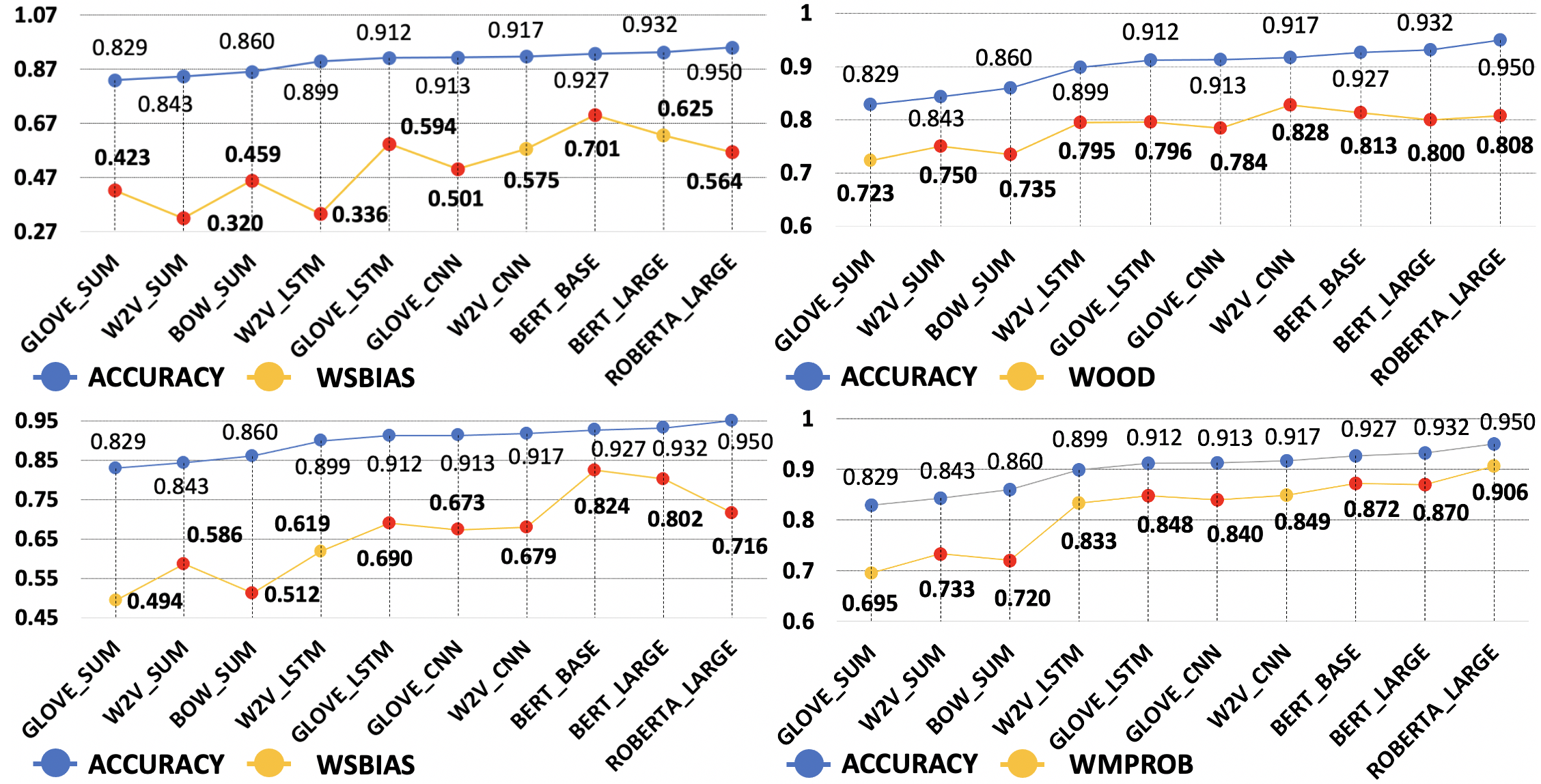}
    \caption{The MLCs shows the model performance based on Accuracy and \textit{WSBias} (from Algorithms \ref{algo:one}, \ref{algo:two}) / \textit{WOOD} / \textit{WMProb}. Here, the yellow dots indicate that the model's ranking position is the same as that of accuracy, red dots indicate that there is a change in ranking position. Weighting scheme considered is Case 1, for 7 equally populated splits on the IMDb dataset.}
    \label{fig:mlc}
\end{figure*}

\noindent \textbf{Least Change in Ranking:}
 The MLC (Figure \ref{fig:mlc}) shows that out of the 3 metrics, accuracy most closely models \textit{WMProb}, with 6/10 models changing ranking position; these changes are local, with models moving up or down a single rank in three swapped pairs.
 
 \noindent\textbf{All Models Perform Poorly While Answering With High Confidence:}
 Unlike the other metrics, here we see that all models exhibit a sharp decrease in split-wise performance in the last split, where prediction confidence is highest.
 
 \noindent\textbf{Accuracy Least Inflates Model Performance in the Context of Prediction Confidence:}
 There is only a ~5\%-16\% decrease in \textit{WMProb} values when compared to accuracy, indicating that while accuracy inflates performance measurement by ignoring model confidence, the influence of model confidence is lesser compared to OOD characteristics and bias used in the other metrics. Similar results are observed in other weight/split considerations.
 
 \section*{Discussion}

We find that GLOVE-LSTM efficiently utilizes spurious bias to solve `easy' questions. Also, W2V-CNN has the highest capability for OOD generalization. While ROBERTA-LARGE has the highest number of correct predictions per split, it more or less uniformly distributes incorrect predictions across all splits (i.e., varying levels of confidence). Therefore, if ROBERTA-LARGE is augmented with the capability seen in BERT-BASE, GLOVE-LSTM and W2V-LSTM of better correlation of prediction correctness with prediction confidence, then it may be better suited for safety critical applications. Additionally, our results over both datasets show that transformers, in general, fail to solve

\noindent `easy' questions; the fact that simpler models solve easy questions more efficiently can be utilized to create ensembles, and subsequently, better architectures. Conducting further experiments on different datasets and tasks is a potential future work to generalize these findings. 

\section*{Prototype for Leaderboard Probing and Customization}
We develop an interactive tool prototype\footnote{https://github.com/swarooprm/Leaderboard-Customization} -- leveraging \textit{WSBias}, \textit{WOOD}, and \textit{WMProb}-- to help industry experts probe and customize leaderboards, as per their deployment scenario requirements. The provision of a platform for such structured analysis is aimed at helping experts select better models, as well as reducing their pre-deployment development and testing effort. We evaluate this prototype in two stages-- (i) Expert Review, and (ii) Detailed User Study.

\subsection{Expert Review}
 We present an initial schematic, covering ten models over the SST-2 and IMDb datasets, to five software testing experts (6+ years of work experience), each working in one of the focus areas illustrated in Figure \ref{introfig}. There is a shared belief among experts that \textit{``the use of this tool will reduce testing time/effort by reducing the number of models initially vetted"} \textbf{(P1, P4)}. Additionally, \textit{``... selecting good base models reduces testing times downstream the pipeline"} \textbf{(P3)}. Based on this initial feedback, we add a feature to custom-load model results and data (as present in conventional software testing platforms). This extends the utility of our prototype from the initial phase of probing and model selection, to extensive, iterative behavioral testing that occurs in later testing phases, prior to deployment.
 
 \subsection{User Study}

\begin{figure*}[t]
    \centering
    \includegraphics[width=0.87\textwidth]{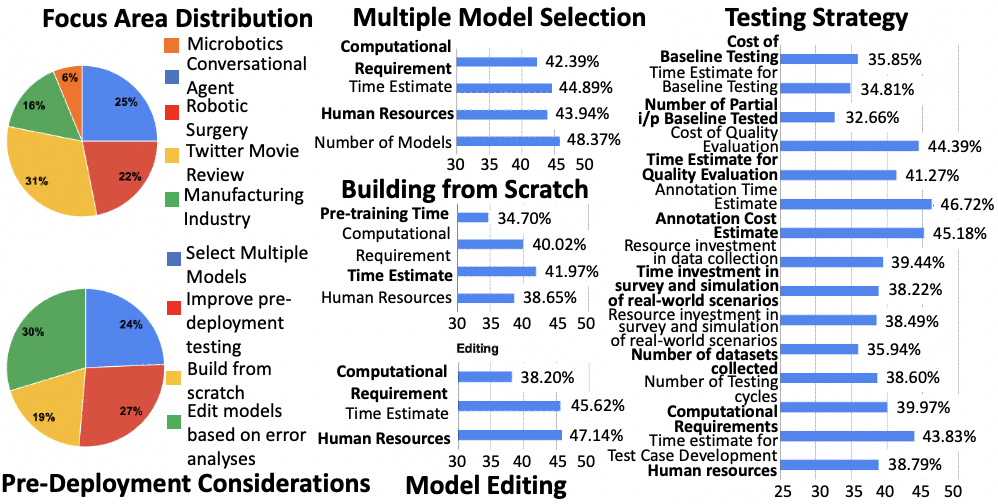}
    \caption{Reported \textbf{Decreases} in pre-deployment development and testing. Average : \textbf{\textit{40.791\%}}}
    \label{fig:results}
\end{figure*}

 We approach software developers and testing managers of several commercial product development teams (belonging to 9 companies) across all five focus areas. We first outline the basic intuitions behind the formulation of the three metrics, and then provide users with our prototype. In the visual interface, users can interactively test various splits (manual/random/automatic methods-- i.e., equally populated/thresholded splits) and weighting schemes (with flexible penalty assignment, and discrete/continuous weighting) of data for different metrics. The users are then asked to fill a Google Form, juxtaposing their prototype usage with conventional model selection-- refers to picking models from leaderboards (with accuracy as the metric, i.e., uniform weighting, no penalty). In conventional selection, users have to iteratively pick some models from the leaderboard, then test those models in their specific application (through test cases and deployment), until they find one that works efficiently. Figure \ref{fig:results} summarizes user study results.

\subsection{Feedback}
A total of 32 experts participated in the evaluation of our prototype. On average, \textbf{development and testing effort decrease of ~41\%} is reported using our customized leaderboard approach, as opposed to conventional model selection and testing approaches. This decrease is attributed to the prototype's \textit{``... improvement for data collection time frames and downstream testing since we know exactly where the model fails."}\textbf{(P7)}. \textbf{(P6)} believes our prototype \textit{``can help in creating model ensembles"}, and \textbf{(P8)} that it will \textit{``help with the expedited development and testing of AI models... auto-suggested model edits through the visual interface would be interesting"}. \textbf{(P14, P31)} say our prototype \textit{... helps compare in-house and commercial model strengths and weaknesses}. \textbf{(P20)} further attributes this facilitated comparison towards enabling \textit{``...testing of various possible scenarios in deployment without writing test cases... we do not have to write separate test cases for stress testing"}, and \textbf{(P26)} adds that \textit{``A big part of testing involves creating new datasets in the sandbox that mimic deployment scenarios. Dataset creation, annotation, controlling quality, and evaluation are costly. This tool's unique method of annotating existing datasets helps us significantly reduce previously spent effort."}. \textbf{(P22)} suggests \textit{``... exploring the potential expansion of this tool to assist in customized functional testing of products"}.

\section*{Related Works}

\textbf{Beyond Accuracy:}
CheckList, a matrix of general linguistic capabilities and test types, has been introduced to enable better model evaluation and address the performance overestimation associated with accuracy \cite{Ribeiro2020BeyondAB}. Alternate evaluation approaches such as model robustness to noise \cite{belinkov2017synthetic}, Perturbation Sensitivity Analysis--a generic evaluation framework detects unintended model biases related to named entities-- \cite{prabhakaran2019perturbation}, and energy usage in models \cite{henderson2020towards} (promoting Green AI \cite{schwartz2019green,Mishra2020DoWN}) have also been proposed. Our proposed idea of weighting samples based on `hardness', is metric-independent, and can be used to revamp other metrics beyond accuracy such as F1 Score.

\noindent \textbf{Evaluation Metrics and Recommendations:} The misalignment of automatic metrics with the gold standard of human evaluation for Text Generation has been studied in several works \cite{Mathur2020TangledUI, Sellam2020BLEURTLR}. Our work is orthogonal to these, as we have no reference human evaluation, and our idea of weighted metrics is task-independent. Evaluation metric analysis has been used to provide recommendations \cite{Peyrard2019StudyingSE} for gathering human annotations in specified scoring ranges \cite{radev2003evaluation}. `Best practices' for reporting experimental results \cite{Dodge2019ShowYW}, the use of randomly generated splits \cite{gorman2019we}, and comparing score distributions on multiple executions \cite{Reimers2017ReportingSD} have been recommended. Our work recommends the reporting of performance in terms of the weighted metrics, along with accuracy, and their integration as part of leaderboard customization.



\noindent \textbf{Adversarial Evaluation:}
Adversarial evaluation tests the capability of systems to not get distracted by text added to samples in several different forms \cite{Jia2017AdversarialEF, jin2019bert, iyyer2018adversarial}, intended to mislead models \cite{Jia2019CertifiedRT}. Our work, in contrast, deals with the adversarial attack of leaderboards. Ladder \cite{blum2015ladder} has been proposed as a reliable leaderboard for machine learning competitions, by preventing participants from overfitting to holdout sets. Our objective is orthogonal to this as it focuses on customizing leaderboards based on application specific requirements in industry.

\section*{Conclusion}

We present a method to adversarially attack leaderboards, and utilize this to probe model rankings. We find that `better' models are not always ranked higher, and therefore propose new evaluation metrics that enable equitable model evaluation, by weighting samples based on their hardness from perspectives of data and model dependencies. We use these metrics to develop a tool prototype for leaderboard customization by adapting software engineering concepts. We perform a user study with experts from nine industries, having five different focus areas, and find that our prototype helps in calibrating the proposed metrics by reducing user development and testing effort on average by ~41\%. Our metrics reduce inflation in model performance, thus rectifying overestimated capabilities of AI systems. Our results also give preliminary indications of the strengths and weaknesses of 10 models. We recommend the use of appropriate evaluation metrics to do fair model evaluation, thus minimizing the gap between research and the real world. In the future, we will expand our analysis to model ensembles, as they are seen to dominate leaderboards more often. Our equitable evaluation metrics can also be useful in competitions, and we plan to expand our user study to competitions in future. We hope the community expands the usage of such evaluation metrics to other domains, such as vision and speech.

\section*{Ethics Statement}

The broad impact of our work is as follows:\\

\noindent\textbf{Environmental Impact}: Competitions and leaderboards in general can have a negative impact on climate change, as increasingly complex models are trained and retrained, requiring more computation time. Our proposed method could help teams reduce the exploration space to find a good model, and thus help reduce that team's overall carbon emissions. 

\noindent\textbf{Towards a Fair Leaderboard}:
Our definition of `difficulty' can be extended based on several different perspectives. A viable application of this is a `fair' customized leaderboard, where models having higher gender/racial bias will be heavily penalized, preventing them from dominating leaderboards.

\noindent\textbf{Selecting the Best Natural Language Understanding (NLU) Systems}: NLU requires a variety of linguistic capabilities and reasoning skills. Incorporating these requirements systematically as `difficulties' in our framework will enable better selection of top NLU systems.

\bibliography{aa}

\section{Organization}
\begin{itemize}
\item More Split-Wise Results
\item More Ranking Changes for Various Hyper-Parameter Considerations
\item Visual Interface for Tool
\item Google Form for User Study
\end{itemize}

\section{More Split-Wise Results}

\begin{figure*}[t]
    \centering
    \includegraphics[width=\textwidth]{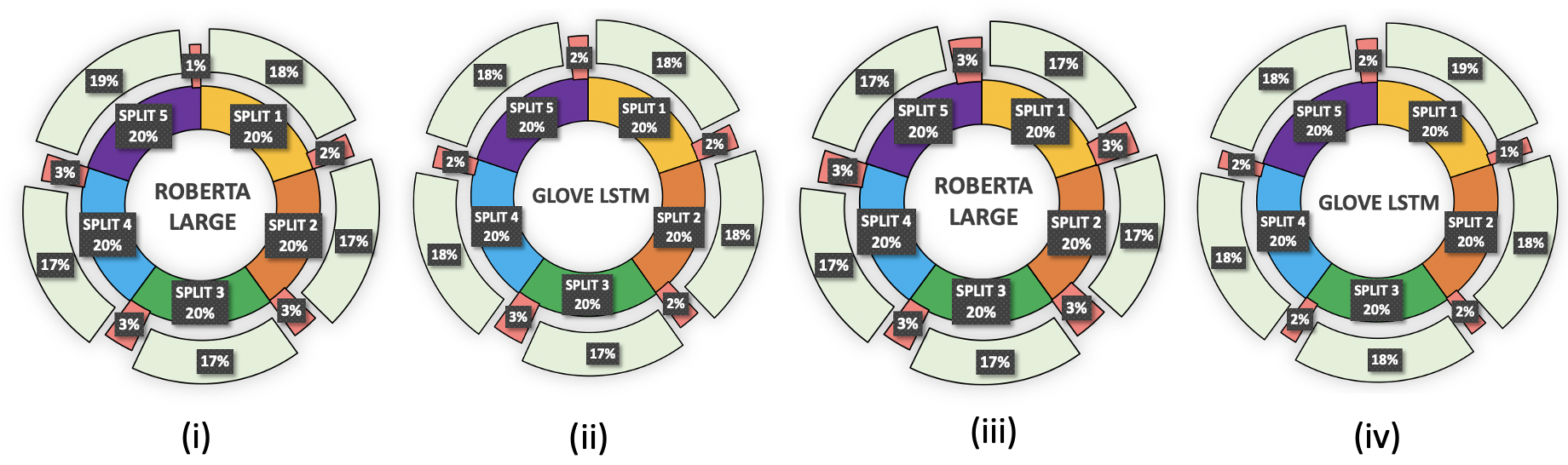}
    \caption{The inner-ring shows how 5 splits are formed in ROBERTA-LARGE and GLOVE-LSTM on the basis of \textit{thresholding} at bias score values (calculated with Algorithm 1) of 0.2, 0.4, 0.6, 0.8 for the IMDb (i,ii) and SST-2 (iii,iv) Datasets. The outer-ring shows the proportion of correct (light-green) /incorrect (light-red) sample predictions per split.}
    \label{f1}
\end{figure*}

\begin{figure*}[t]
    \centering
    \includegraphics[width=0.8\textwidth]{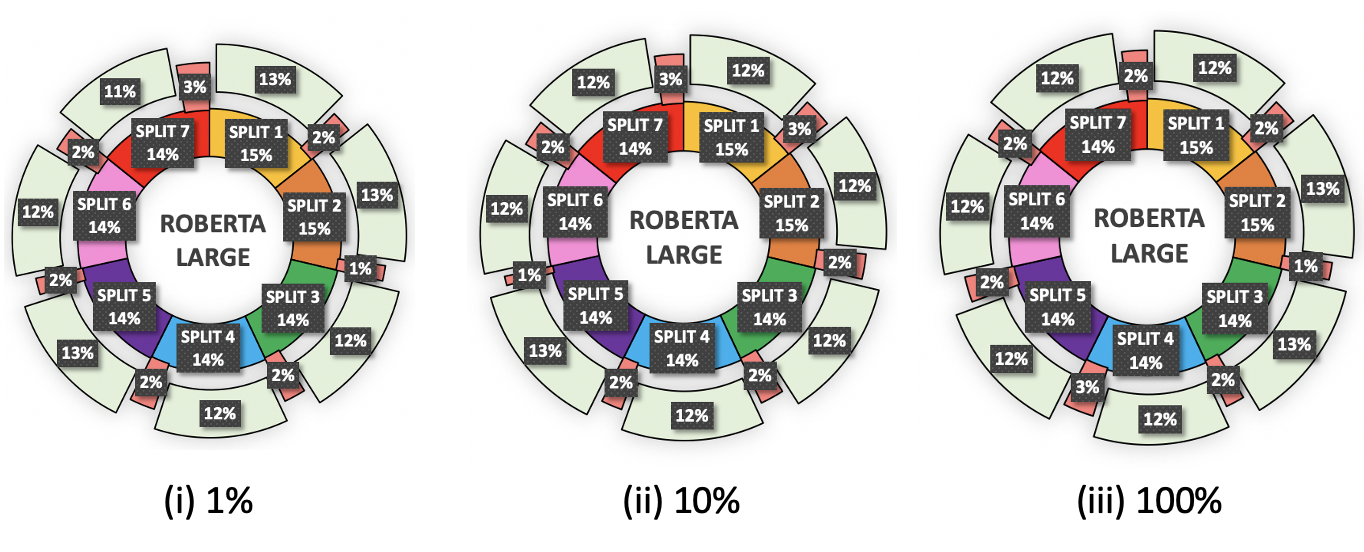}
    \caption{The inner-ring shows how 7 equal splits are formed in ROBERTA-LARGE based on ranking using averaged STS values of test set samples with respect to the Top 1\% (i) ,10\% (ii) ,100\% (iii) of train set samples for the SST-2 dataset. The outer ring shows the proportion of correct (light-green)/incorrect (light-red) predictions per split.}
    \label{f2}
\end{figure*}

\begin{figure*}[t]
    \centering
    \includegraphics[width=\textwidth]{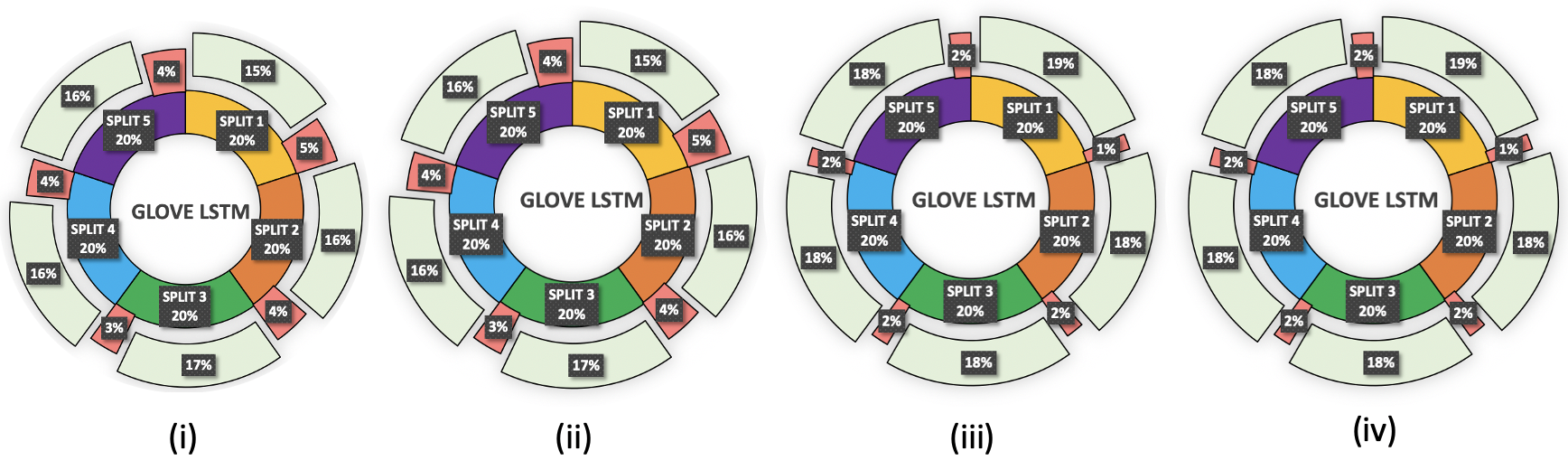}
    \caption{Inner ring shows the sample distribution over 5 splits formed on the basis of: (i,iii) thresholding at averaged confidence values of 0.2, 0.4, 0.6, 0.8, and (ii,iv) equal number of samples per split, for the SST-2 (iii,iv) and IMDb (i,ii) datasets using confidence values of predictions by the GLOVE-LSTM model. The outer-ring shows the proportions of correct (light-green)/incorrect (light-red) samples classified per split.}
    \label{f3}
\end{figure*}

\begin{figure*}[t]
    \centering
    \includegraphics[width=\textwidth]{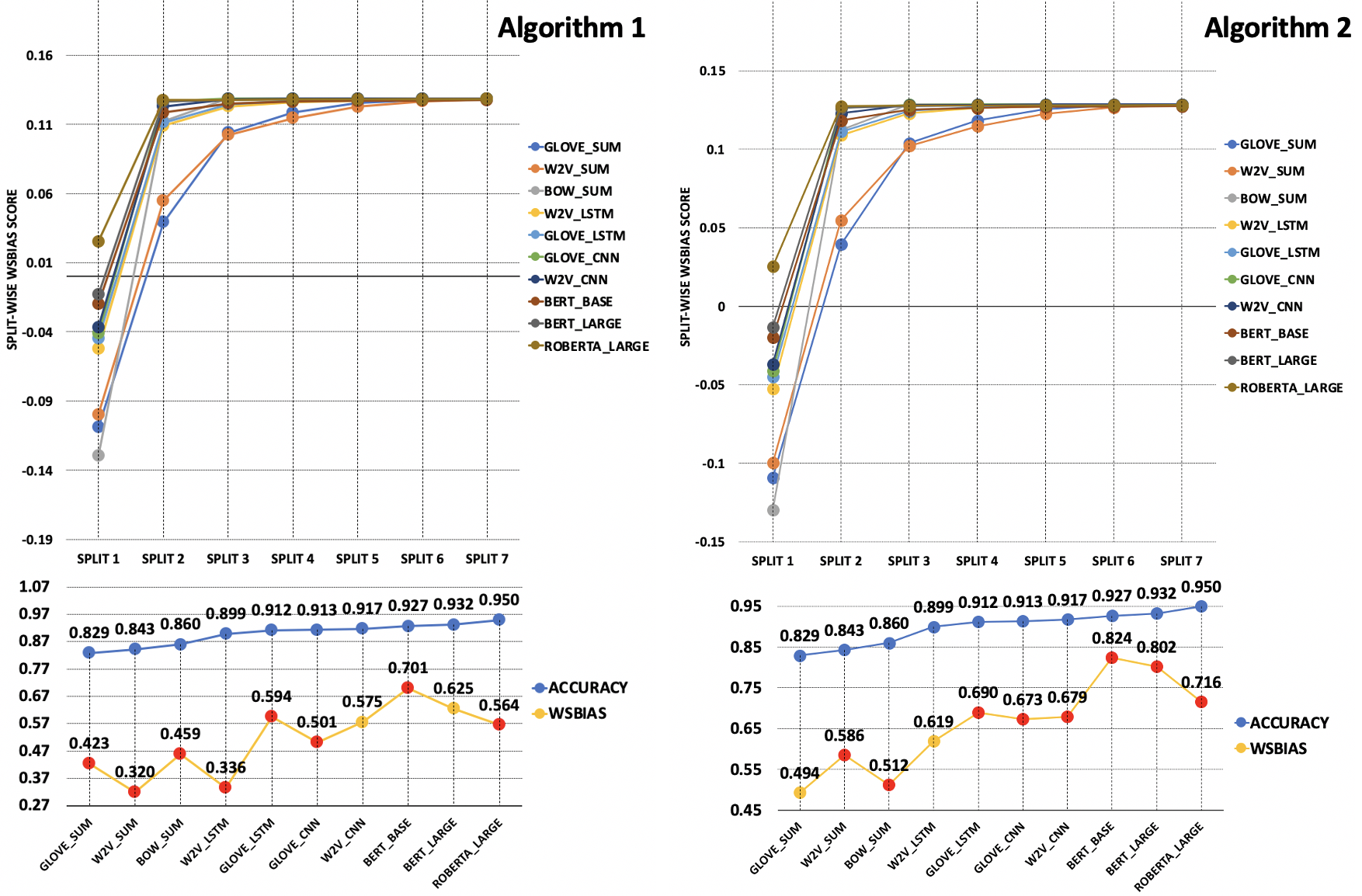}
    \caption{Split-wise results for each model using bias scores from both algorithms are  shown in the parallel coordinates plots (PCP) on the upper half of the figure. Each vertical line indicates the data split considered (Split 1 --Easiest, Split 7 --Hardest). The multi-line charts (MLC) show the model performance based on Accuracy and WSBIAS. Here, on the WSBIAS line, yellow dots indicate that the model's ranking position is the same as that of accuracy, red dots indicate that there is a change in ranking position. Weighting scheme considered is Case 1, for 7 equal splits on the IMDb dataset.}
    \label{f4}
\end{figure*}

Please refer to Figures \ref{f1},\ref{f2},\ref{f3},\ref{f4}.


\section{More  Ranking  Changes  for  Various  Hyper-Parameter Considerations}

Please refer to Tables \ref{t1},\ref{t2},\ref{t3},\ref{t4},\ref{t5},\ref{t6},\ref{t7},\ref{t8}

\definecolor{myRed}{rgb}{0.97,0.52,0.3}
\definecolor{myYellow}{rgb}{0.96,0.84,0.37}
\definecolor{myGreen}{rgb}{0.56,0.93,0.56}

\begin{table*}[t]
\centering
\scriptsize
\resizebox{\textwidth}{!}{%
\begin{tabular}{|c|c|c|c|c|c|c|c|}
\hline
\textbf{Accuracy}                      & \multicolumn{7}{c|}{\textbf{WOOD}}                                                                                                                                                                                                                                                      \\ \hline
\textbf{SST-2} & \textbf{7 Splits}                     & \textbf{6 Splits}                     & \textbf{5 Splits}                       & \textbf{4 Splits}                     & \textbf{3 Splits}                     & \textbf{2 Splits}                     & \textbf{Overall Score}                \\ \hline
BOW SUM                  & \cellcolor{myYellow}W2V SUM          & \cellcolor{myYellow}GLOVE SUM        & \cellcolor{myYellow}GLOVE SUM & \cellcolor{myYellow}GLOVE SUM        & \cellcolor{myYellow}GLOVE SUM        & \cellcolor{myYellow}GLOVE SUM        & \cellcolor{myRed}GLOVE SUM     \\ \hline
GLOVE SUM                    & \cellcolor{white}GLOVE SUM     & \cellcolor{myYellow}W2V SUM          & \cellcolor{myYellow}W2V SUM            & \cellcolor{myYellow}W2V SUM          & \cellcolor{myYellow}W2V SUM          & \cellcolor{myYellow}BOW SUM          & \cellcolor{myRed}W2V SUM       \\ \hline
W2V SUM                      & \cellcolor{myYellow}W2V LSTM         & \cellcolor{myYellow}W2V LSTM         & \cellcolor{myYellow}BOW SUM            & \cellcolor{myYellow}BOW SUM          & \cellcolor{myYellow}BOW SUM          & \cellcolor{white}W2V SUM       & \cellcolor{myRed}BOW SUM       \\ \hline
W2V LSTM                     & \cellcolor{myYellow}GLOVE CNN        & \cellcolor{myYellow}GLOVE CNN        & \cellcolor{white}W2V LSTM        & \cellcolor{white}W2V LSTM      & \cellcolor{white}W2V LSTM      & \cellcolor{white}W2V LSTM      & \cellcolor{myGreen}W2V LSTM      \\ \hline
GLOVE CNN                    & \cellcolor{myYellow}W2V CNN          & \cellcolor{myYellow}BOW SUM          & \cellcolor{white}GLOVE CNN       & \cellcolor{white}GLOVE CNN     & \cellcolor{white}GLOVE CNN     & \cellcolor{myYellow}GLOVE LSTM       & \cellcolor{myRed}GLOVE LSTM    \\ \hline
W2V CNN                      & \cellcolor{myYellow}BOW SUM          & \cellcolor{white}W2V CNN       & \cellcolor{white}W2V CNN         & \cellcolor{white}W2V CNN       & \cellcolor{myYellow}GLOVE LSTM       & \cellcolor{myYellow}GLOVE CNN        & \cellcolor{myRed}GLOVE CNN     \\ \hline
GLOVE LSTM                   & \cellcolor{white}GLOVE LSTM    & \cellcolor{white}GLOVE LSTM    & \cellcolor{white}GLOVE LSTM      & \cellcolor{white}GLOVE LSTM    & \cellcolor{myYellow}W2V CNN          & \cellcolor{myYellow}W2V CNN          & \cellcolor{myRed}W2V CNN       \\ \hline
ROBERTA LARGE                & \cellcolor{white}ROBERTA LARGE & \cellcolor{white}ROBERTA LARGE & \cellcolor{white}ROBERTA LARGE   & \cellcolor{white}ROBERTA LARGE & \cellcolor{white}ROBERTA LARGE & \cellcolor{white}ROBERTA LARGE & \cellcolor{myRed}BERT BASE     \\ \hline
BERT LARGE                   & \cellcolor{myYellow}BERT BASE        & \cellcolor{white}BERT LARGE    & \cellcolor{white}BERT LARGE      & \cellcolor{myYellow}BERT BASE        & \cellcolor{white}BERT LARGE    & \cellcolor{white}BERT LARGE    & \cellcolor{myRed}BERT LARGE    \\ \hline
BERT BASE                    & \cellcolor{myYellow}BERT LARGE       & \cellcolor{white}BERT BASE     & \cellcolor{white}BERT BASE       & \cellcolor{myYellow}BERT LARGE       & \cellcolor{white}BERT BASE     & \cellcolor{white}BERT BASE     & \cellcolor{myRed}ROBERTA LARGE \\ \hline
\end{tabular}%
}
\caption{SST-2 : WOOD Score Results: STS calculated using the top 1\%, 5\%, 10\%, 25\%, 30\%, 40\%, 50\%, 75\%, 100\% of the training data, over equally populated splits. Results are the same for all weight schemes. Orange and red cells represent changes in ranking. White and green cells represent no change in ranking.}
\label{t1}
\end{table*}

\begin{table*}
\scriptsize
\centering
\resizebox{\textwidth}{!}{%
\begin{tabular}{|c|c|c|c|c|c|c|c|}
\hline
\textbf{Accuracy}                  & \multicolumn{7}{c|}{\textbf{WOOD}}                                                                                                                                                                                                                                                       \\ \hline
\cellcolor{white}\textbf{IMDB} & \textbf{7 Splits}                       & \textbf{6 Splits}                       & \textbf{5 Splits}                       & \textbf{4 Splits}                       & \textbf{3 Splits}                  & \textbf{2 Splits}                  & \textbf{Overall Score}                 \\ \hline
GLOVE SUM                          & \cellcolor{myYellow}\textbf{GLOVE SUM} & \cellcolor{myYellow}\textbf{GLOVE SUM} & \cellcolor{myYellow}\textbf{GLOVE SUM} & \cellcolor{myYellow}\textbf{GLOVE SUM} & \cellcolor{myYellow}GLOVE SUM     & \cellcolor{myYellow}GLOVE SUM     & \cellcolor{myGreen}GLOVE SUM      \\ \hline
W2V SUM                            & \cellcolor{myYellow}BOW SUM            & \cellcolor{myYellow}BOW SUM            & \cellcolor{myYellow}BOW SUM            & \cellcolor{myYellow}BOW SUM            & \cellcolor{myYellow}BOW SUM       & \cellcolor{myYellow}W2V SUM       & \cellcolor{myRed}BOW SUM        \\ \hline
BOW SUM                            & \cellcolor{myYellow}W2V SUM            & \cellcolor{myYellow}W2V SUM            & \cellcolor{myYellow}W2V SUM            & \cellcolor{myYellow}W2V SUM            & \cellcolor{myYellow}W2V SUM       & \cellcolor{white}BOW SUM       & \cellcolor{myRed}W2V SUM        \\ \hline
W2V LSTM                           & \cellcolor{myYellow}GLOVE LSTM         & \cellcolor{myYellow}GLOVE LSTM         & \cellcolor{white}W2V LSTM           & \cellcolor{white}W2V LSTM           & \cellcolor{white}W2V LSTM      & \cellcolor{white}W2V LSTM      & \cellcolor{myRed}GLOVE CNN      \\ \hline
GLOVE LSTM                         & \cellcolor{myYellow}GLOVE CNN          & \cellcolor{myYellow}W2V LSTM           & \cellcolor{white}GLOVE LSTM         & \cellcolor{white}GLOVE LSTM         & \cellcolor{myYellow}GLOVE CNN     & \cellcolor{myYellow}GLOVE LSTM    & \cellcolor{myRed}W2V LSTM       \\ \hline
GLOVE CNN                          & \cellcolor{myYellow}W2V LSTM           & \cellcolor{white}GLOVE CNN          & \cellcolor{white}GLOVE CNN          & \cellcolor{white}GLOVE CNN          & \cellcolor{myYellow}GLOVE LSTM    & \cellcolor{myYellow}GLOVE CNN     & \cellcolor{myRed}GLOVE LSTM     \\ \hline
W2V CNN                            & \cellcolor{myYellow}ROBERTA LARGE      & \cellcolor{myYellow}BERT LARGE         & \cellcolor{white}W2V CNN            & \cellcolor{myYellow}BERT LARGE         & \cellcolor{myYellow}W2V CNN       & \cellcolor{myYellow}W2V CNN       & \cellcolor{myRed}BERT LARGE     \\ \hline
BERT BASE                          & \cellcolor{myYellow}BERT LARGE         & \cellcolor{myYellow}ROBERTA LARGE      & \cellcolor{myYellow}BERT LARGE         & \cellcolor{myYellow}W2V CNN            & \cellcolor{myYellow}BERT LARGE    & \cellcolor{myYellow}BERT LARGE    & \cellcolor{myRed}ROBERTA\_LARGE \\ \hline
BERT LARGE                         & \cellcolor{myYellow}W2V CNN            & \cellcolor{myYellow}W2V CNN            & \cellcolor{myYellow}ROBERTA LARGE      & \cellcolor{myYellow}ROBERTA LARGE      & \cellcolor{myYellow}ROBERTA LARGE & \cellcolor{myYellow}BERT BASE     & \cellcolor{myRed}BERT BASE      \\ \hline
ROBERTA LARGE                      & \cellcolor{myYellow}BERT BASE          & \cellcolor{myYellow}BERT BASE          & \cellcolor{myYellow}BERT BASE          & \cellcolor{myYellow}BERT BASE          & \cellcolor{myYellow}BERT BASE     & \cellcolor{white}ROBERTA LARGE & \cellcolor{myRed}W2V CNN        \\ \hline
\end{tabular}%
}
\caption{IMDb : WOOD Score Results: STS calculated using the top 1\%, 5\%, 10\%, 25\%, 30\%, 40\%, 50\%, 75\%, 100\% of the training data, over equally populated splits. Results are the same for all weight schemes. Orange and red cells represent changes in ranking. White and green cells represent no change in ranking.}
\label{t2}
\end{table*}

\begin{table*}
\scriptsize
\centering
\resizebox{\textwidth}{!}{%
\begin{tabular}{|c|l|l|l|l|}
\hline
\textbf{Accuracy}                      & \multicolumn{4}{c|}{\textbf{WOOD}}                                                                                                                                    \\ \hline
\textbf{SST-2} & \multicolumn{1}{c|}{\textbf{5 Splits}}  & \multicolumn{1}{c|}{\textbf{4 Splits}}  & \multicolumn{1}{c|}{\textbf{3 Splits}}  & \multicolumn{1}{c|}{\textbf{2 Splits}}  \\ \hline
BOW SUM                               & \cellcolor{myYellow}GLOVE SUM & \cellcolor{myYellow}GLOVE SUM & \cellcolor{myYellow}GLOVE SUM & \cellcolor{myYellow}GLOVE SUM \\ \hline
GLOVE SUM                             & \cellcolor{myYellow}W2V SUM            & \cellcolor{myYellow}W2V SUM            & \cellcolor{myYellow}W2V SUM            & \cellcolor{myYellow}BOW SUM            \\ \hline
W2V SUM                               & \cellcolor{myYellow}BOW SUM            & \cellcolor{myYellow}BOW SUM            & \cellcolor{myYellow}BOW SUM            & \cellcolor{white}W2V SUM            \\ \hline
W2V LSTM                              & \cellcolor{white}W2V LSTM           & \cellcolor{white}W2V LSTM           & \cellcolor{white}W2V LSTM           & \cellcolor{white}W2V LSTM           \\ \hline
GLOVE CNN                             & \cellcolor{white}GLOVE CNN          & \cellcolor{white}GLOVE CNN          & \cellcolor{white}GLOVE CNN          & \cellcolor{myYellow}GLOVE LSTM         \\ \hline
W2V CNN                               & \cellcolor{white}W2V CNN            & \cellcolor{white}W2V CNN            & \cellcolor{myYellow}GLOVE LSTM         & \cellcolor{myYellow}GLOVE CNN          \\ \hline
GLOVE LSTM                            & \cellcolor{white}GLOVE LSTM         & \cellcolor{white}GLOVE LSTM         & \cellcolor{myYellow}W2V CNN            & \cellcolor{myYellow}W2V CNN            \\ \hline
ROBERTA LARGE                         & \cellcolor{white}ROBERTA LARGE      & \cellcolor{white}ROBERTA LARGE      & \cellcolor{white}ROBERTA LARGE      & \cellcolor{white}ROBERTA LARGE      \\ \hline
BERT LARGE                            & \cellcolor{white}BERT LARGE         & \cellcolor{myYellow}BERT BASE          & \cellcolor{white}BERT LARGE         & \cellcolor{white}BERT LARGE         \\ \hline
BERT BASE                             & \cellcolor{white}BERT BASE          & \cellcolor{myYellow}BERT LARGE         & \cellcolor{white}BERT BASE          & \cellcolor{white}BERT BASE          \\ \hline
\end{tabular}%
}
\caption{SST-2 : WOOD Score Results: STS calculated using the top 1\%, 5\%, 10\%, 25\%, 30\%, 40\%, 50\%, 75\%, 100\% of the training data, over equally spaced (between 0-1) thresholded splits. Results are the same for all weight schemes. Orange cells represent changes in ranking. White cells represent no change in ranking.}
\label{t3}
\end{table*}

\begin{table*}
\centering
\scriptsize
\resizebox{\textwidth}{!}{%
\begin{tabular}{|c|c|c|c|c|}
\hline
\textbf{Accuracy}                  & \multicolumn{4}{c|}{\textbf{WOOD}}                                                                                                             \\ \hline
\textbf{IMDB} & \textbf{5 Splits}                  & \textbf{4 Splits}                  & \textbf{3 Splits}                  & \textbf{2 Splits}               \\ \hline
GLOVE SUM                          & GLOVE SUM                          & GLOVE SUM                          & GLOVE SUM                          & GLOVE SUM                       \\ \hline
W2V SUM                            & \cellcolor{myYellow}BOW SUM       & \cellcolor{myYellow}BOW SUM       & \cellcolor{myYellow}BOW SUM       & W2V SUM                         \\ \hline
BOW SUM                            & \cellcolor{myYellow}W2V SUM       & \cellcolor{myYellow}W2V SUM       & \cellcolor{myYellow}W2V SUM       & BOW SUM                         \\ \hline
W2V LSTM                           & W2V LSTM                           & W2V LSTM                           & W2V LSTM                           & W2V LSTM                        \\ \hline
GLOVE LSTM                         & GLOVE LSTM                         & GLOVE LSTM                         & \cellcolor{myYellow}GLOVE CNN     & GLOVE LSTM                      \\ \hline
GLOVE CNN                          & GLOVE CNN                          & GLOVE CNN                          & \cellcolor{myYellow}GLOVE LSTM    & GLOVE CNN                       \\ \hline
W2V CNN                            & W2V CNN                            & \cellcolor{myYellow}BERT LARGE    & W2V CNN                            & W2V CNN                         \\ \hline
BERT BASE                          & \cellcolor{myYellow}BERT LARGE    & \cellcolor{myYellow}W2V CNN       & \cellcolor{myYellow}BERT LARGE    & \cellcolor{myYellow}BERT LARGE \\ \hline
BERT LARGE                         & \cellcolor{myYellow}ROBERTA LARGE & \cellcolor{myYellow}ROBERTA LARGE & \cellcolor{myYellow}ROBERTA LARGE & \cellcolor{myYellow}BERT BASE  \\ \hline
ROBERTA LARGE                      & \cellcolor{myYellow}BERT BASE     & \cellcolor{myYellow}BERT BASE     & \cellcolor{myYellow}BERT BASE     & ROBERTA LARGE                   \\ \hline
\end{tabular}%
}
\caption{IMDb : WOOD Score Results: STS calculated using the top 1\%, 5\%, 10\%, 25\%, 30\%, 40\%, 50\%, 75\%, 100\% of the training data, over equally spaced (between 0-1) thresholded splits. Results are the same for all weight schemes. Orange cells represent changes in ranking. White cells represent no change in ranking.}
\label{t4}
\end{table*}

\begin{table*}
\scriptsize
\centering
\resizebox{\textwidth}{!}{%
\begin{tabular}{|c|c|c|c|c|c|c|}
\hline
\textbf{Accuracy}                      & \multicolumn{6}{c|}{\textbf{WSBIAS}}                                                                                                                                                                                 \\ \hline
\textbf{SST-2} & \textbf{ Algo 1 Regular Splits}     & \textbf{ Algo 2 Regular Splits} & \textbf{ Algo 1 Continuous Weights} & \textbf{ Algo 2 Continuous Weights} & \textbf{ Algo 1 Threshold Splits}   & \textbf{ Algo 2 Threshold Splits} \\ \hline
BOW SUM                      & \cellcolor {myYellow}W2V-SUM       & \cellcolor {myYellow}GLOVE-SUM & \cellcolor {myYellow}GLOVE-SUM     & \cellcolor {myYellow}GLOVE-SUM    & \cellcolor {myYellow}W2V-SUM       & \cellcolor {myYellow}GLOVE-SUM   \\ \hline
GLOVE SUM                    & \cellcolor {myYellow}W2V-LSTM      & \cellcolor {myYellow}BOW-SUM   & \cellcolor {myYellow}W2V-LSTM      & \cellcolor {myYellow}W2V-LSTM     & \cellcolor {myYellow}W2V-LSTM      & \cellcolor {myYellow}BOW-SUM     \\ \hline
W2V SUM                      & \cellcolor {myYellow}GLOVE-SUM     & W2V-SUM                        & \cellcolor {myYellow}BOW-SUM       & \cellcolor {myYellow}BOW-SUM      & \cellcolor {myYellow}GLOVE-SUM     & W2V-SUM                          \\ \hline
W2V LSTM                     & \cellcolor {myYellow}BOW-SUM       & W2V-LSTM                       & \cellcolor {myYellow}W2V-SUM       & \cellcolor {myYellow}W2V-SUM      & \cellcolor {myYellow}BOW-SUM       & W2V-LSTM                         \\ \hline
GLOVE CNN                    & GLOVE-CNN                          & GLOVE-CNN                      & GLOVE-CNN                          & GLOVE-CNN                         & GLOVE-CNN                          & GLOVE-CNN                        \\ \hline
W2V CNN                      & \cellcolor {myYellow}ROBERTA-LARGE & W2V-CNN                        & W2V CNN                            & W2V CNN                           & \cellcolor {myYellow}ROBERTA-LARGE & W2V-CNN                          \\ \hline
GLOVE LSTM                   & \cellcolor {myYellow}W2V-CNN       & GLOVE-LSTM                     & GLOVE LSTM                         & GLOVE LSTM                        & \cellcolor {myYellow}W2V-CNN       & GLOVE-LSTM                       \\ \hline
ROBERTA LARGE                & \cellcolor {myYellow}GLOVE-LSTM    & ROBERTA-LARGE                  & ROBERTA LARGE                      & ROBERTA LARGE                     & \cellcolor {myYellow}GLOVE-LSTM    & ROBERTA-LARGE                    \\ \hline
BERT LARGE                   & BERT-LARGE                         & BERT-LARGE                     & BERT LARGE                         & BERT LARGE                        & BERT-LARGE                         & BERT-LARGE                       \\ \hline
BERT BASE                    & BERT-BASE                          & BERT-BASE                      & BERT BASE                          & BERT BASE                         & BERT-BASE                          & BERT-BASE                        \\ \hline
\end{tabular}%
}
\caption{SST-2 : WSBIAS Score Results: Calculated using bias scores over 7,6,5,4,3,2 splits and overall score. Results are the same for all weight schemes. Orange cells represent changes in ranking. White cells represent no change in ranking.}
\label{t5}
\end{table*}

\begin{table*}
\centering
\scriptsize
\resizebox{\textwidth}{!}{%
\begin{tabular}{|
>{\columncolor{white}}c |
>{\columncolor{myYellow}}c |
>{\columncolor{myYellow}}c |
>{\columncolor{white}}c |
>{\columncolor{white}}c |
>{\columncolor{myYellow}}c |
>{\columncolor{myYellow}}c |}
\hline
\textbf{Accuracy} & \multicolumn{6}{c|}{\cellcolor{white}\textbf{WSBIAS}}\\ \hline
\textbf{IMDB}     & \cellcolor{white}\textbf{Algo 1 Regular Splits} & \cellcolor{white}\textbf{Algo 2 Regular Splits} & \cellcolor{white}\textbf{Algo 1 Continuous Weights} & \cellcolor{white}\textbf{Algo 2 Continuous Weights} & \cellcolor{white}\textbf{Algo 1 Threshold Splits} & \cellcolor{white}\textbf{Algo 2 Threshold Splits} \\ \hline
GLOVE SUM         & W2V-SUM                                                & \cellcolor{white}GLOVE SUM                      & GLOVE SUM                                                  & GLOVE SUM                                                 & W2V-SUM                                                  & \cellcolor{white}GLOVE SUM                        \\ \hline
W2V SUM           & W2V-LSTM                                               & BOW SUM                                                & W2V SUM                                                    & W2V SUM                                                   & W2V-LSTM                                                 & BOW SUM                                                  \\ \hline
BOW SUM           & GLOVE-SUM                                              & W2V SUM                                                & BOW SUM                                                    & BOW SUM                                                   & GLOVE-SUM                                                & W2V SUM                                                  \\ \hline
W2V LSTM          & BOW-SUM                                                & \cellcolor{white}W2V LSTM                       & W2V LSTM                                                   & W2V LSTM                                                  & BOW-SUM                                                  & \cellcolor{white}W2V LSTM                         \\ \hline
GLOVE LSTM        & GLOVE-CNN                                              & GLOVE CNN                                              & GLOVE LSTM                                                 & GLOVE LSTM                                                & GLOVE-CNN                                                & GLOVE CNN                                                \\ \hline
GLOVE CNN         & ROBERTA-LARGE                                          & W2V CNN                                                & GLOVE CNN                                                  & GLOVE CNN                                                 & ROBERTA-LARGE                                            & W2V CNN                                                  \\ \hline
W2V CNN           & \cellcolor{white}W2V-CNN                        & GLOVE LSTM                                             & W2V CNN                                                    & W2V CNN                                                   & \cellcolor{white}W2V-CNN                          & GLOVE LSTM                                               \\ \hline
BERT BASE         & GLOVE-LSTM                                             & ROBERTA LARGE                                          & BERT BASE                                                  & BERT BASE                                                 & GLOVE-LSTM                                               & ROBERTA LARGE                                            \\ \hline
BERT LARGE        & \cellcolor{white}BERT-LARGE                     & BERT LARGE                                             & BERT LARGE                                                 & BERT LARGE                                                & \cellcolor{white}BERT-LARGE                       & BERT LARGE                                               \\ \hline
ROBERTA LARGE     & BERT-BASE                                              & BERT BASE                                              & ROBERTA LARGE                                              & ROBERTA LARGE                                             & BERT-BASE                                                & BERT BASE                                                \\ \hline
\end{tabular}%
}
\caption{IMDb : WSBIAS Score Results: Calculated using bias scores over 7,6,5,4,3,2 splits and overall score. Results are the same for all weight schemes. Orange cells represent changes in ranking. White cells represent no change in ranking.}
\label{t6}
\end{table*}

\begin{table*}
\centering
\scriptsize
\resizebox{0.5\textwidth}{!}{%
\begin{tabular}{|c|c|c|}
\hline
\textbf{Accuracy}                      & \multicolumn{2}{c|}{\textbf{WMPROB}}                       \\ \hline
\textbf{SST-2} & \textbf{Equal Splits}          & \textbf{Threshold Splits} \\ \hline
BOW SUM                               & \cellcolor{myYellow}GLOVE-SUM & BOW SUM                   \\ \hline
GLOVE SUM                             & \cellcolor{myYellow}BOW-SUM   & GLOVE SUM                 \\ \hline
W2V SUM                               & W2V-SUM                        & W2V SUM                   \\ \hline
W2V LSTM                              & W2V-LSTM                       & W2V LSTM                  \\ \hline
GLOVE CNN                             & GLOVE CNN                      & GLOVE CNN                 \\ \hline
W2V CNN                               & W2V CNN                        & W2V CNN                   \\ \hline
GLOVE LSTM                            & GLOVE LSTM                     & GLOVE LSTM                \\ \hline
ROBERTA LARGE                         & ROBERTA LARGE                  & ROBERTA LARGE             \\ \hline
BERT LARGE                            & BERT LARGE                     & BERT LARGE                \\ \hline
BERT BASE                             & BERT BASE                      & BERT BASE                 \\ \hline
\end{tabular}%
}
\caption{SST-2 : WMPROB Score Results: Over 7,6,5,4,3,2 splits and overall score. Results are the same for all weight schemes. Orange cells represent changes in ranking. White cells represent no change in ranking.}
\label{t7}
\end{table*}

\begin{table*}
\centering
\scriptsize
\resizebox{0.5\textwidth}{!}{%
\begin{tabular}{|c|c|c|}
\hline
\textbf{Accuracy}                  & \multicolumn{2}{c|}{\textbf{WMPROB}}                                    \\ \hline
\cellcolor{white}\textbf{IMDB} & \textbf{Equal Splits}              & \textbf{Threshold Splits}          \\ \hline
GLOVE SUM                          & \cellcolor{white}GLOVE-SUM     & \cellcolor{white}GLOVE-SUM     \\ \hline
W2V SUM                            & \cellcolor{white}W2V-SUM       & \cellcolor{myYellow}BOW-SUM       \\ \hline
BOW SUM                            & \cellcolor{white}BOW-SUM       & \cellcolor{myYellow}W2V SUM       \\ \hline
W2V LSTM                           & \cellcolor{white}W2V-LSTM   & \cellcolor{white}W2V-LSTM   \\ \hline
GLOVE LSTM                         & \cellcolor{myYellow}GLOVE-CNN     & \cellcolor{myYellow}GLOVE-CNN     \\ \hline
GLOVE CNN                          & \cellcolor{myYellow}W2V-CNN       & \cellcolor{myYellow}GLOVE-LSTM    \\ \hline
W2V CNN                            & \cellcolor{myYellow}GLOVE-LSTM    & \cellcolor{white}W2V CNN    \\ \hline
BERT BASE                          & \cellcolor{myYellow}BERT-LARGE    & \cellcolor{myYellow}BERT-LARGE    \\ \hline
BERT LARGE                         & \cellcolor{myYellow}BERT-BASE     & \cellcolor{myYellow}BERT-BASE     \\ \hline
ROBERTA LARGE                      & \cellcolor{white}ROBERTA-LARGE & \cellcolor{white}ROBERTA-LARGE \\ \hline
\end{tabular}%
}
\caption{IMDb : WMPROB Score Results:Over 7,6,5,4,3,2 splits and overall score. Results are the same for all weight schemes. Orange cells represent changes in ranking. White cells represent no change in ranking.}
\label{t8}
\end{table*}

\section{Visual Interface for Tool}

\begin{figure*}[t]
    \centering
    \includegraphics[width=\textwidth]{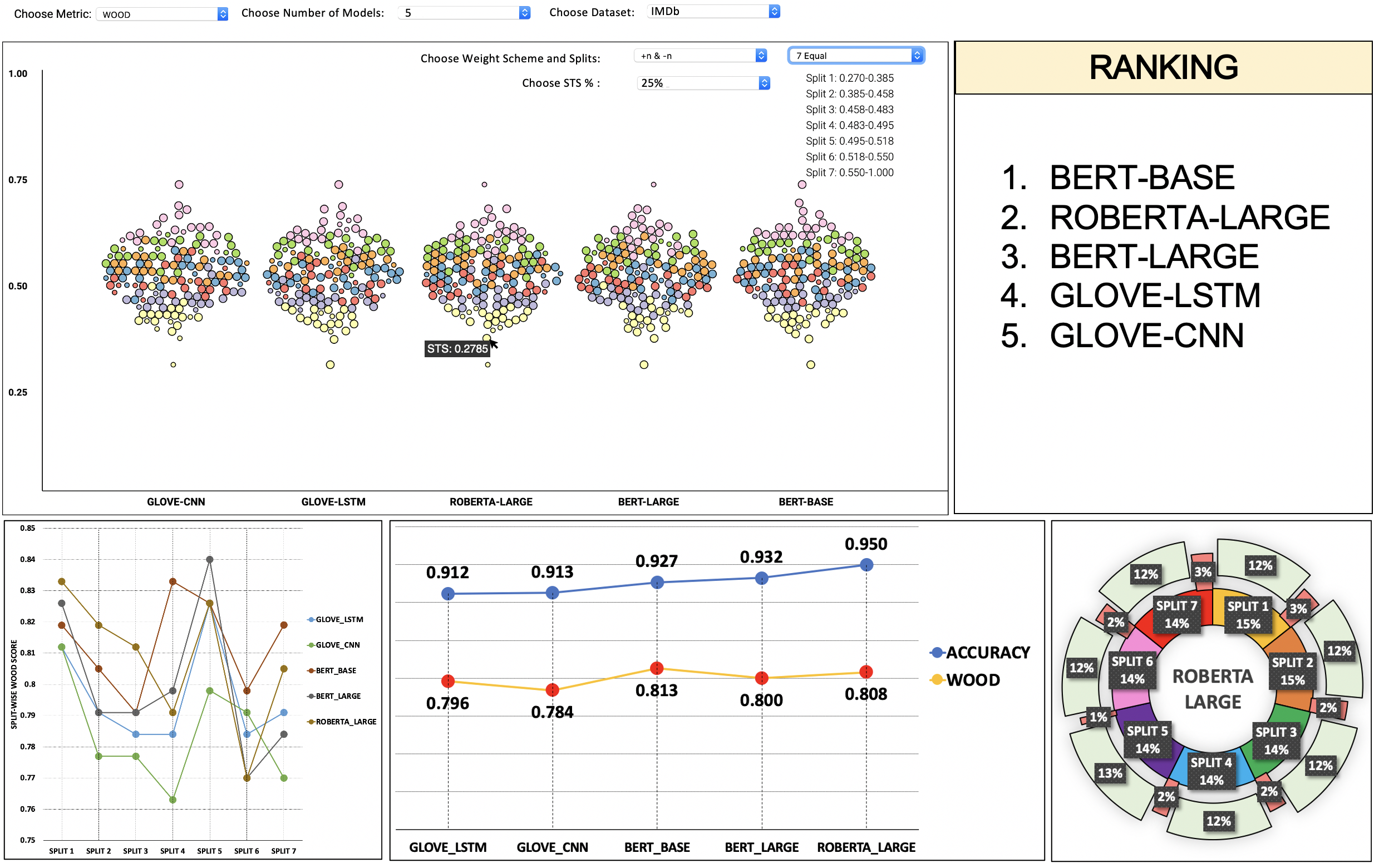}
    \caption{Visual Interface used in User Study: The user selects the metric, number of models to analyze, dataset, weight scheme, \% of train samples to calculate average STS (if metric is WOOD), and the number (and basis--continuous, thresholded, equally populated, manually selected) splits. Based on this, a series of parallel beeswarm plots are generated. x-axis comprises of the models used, and y-axis of the `difficulty' (here, based on STS) of the samples (so sample points at higher y-values are easy samples). The coloring is based on the split the sample belongs to, and the size indicates if the sample is correctly(larger) or incorrectly (smaller) classified. A PCP and MLC for the points is automatically generated. On hover over a sample, based on the model (i.e., corresponding x-coordinate), the sunburst chart (bottom right) for the model is generated. Finally, the new ranking based on the metric selected is displayed.}
    \label{i1}
\end{figure*}

Please refer to Figure \ref{i1}.

\section{Google Form for User Study}
The study covers cases where users find model performance in real world scenarios to range from 1-8 (/10). The user spread over the five application types is also balanced. Users are specifically asked to initially select a product(s) closest to their line of work from the following options:
(i) Conversational AI (Alexa/Siri): OOD data is frequent and inevitable due to low control over user input, (ii) Robotic Surgery: Safety critical applications cannot afford incorrect model answering, and require models to abstain if they are unable to answer, (iii) Twitter Movie Review Classification (Sentiment Analysis): Spurious Bias may play a significant role in fooling a model due to the frequent repetition of certain words or patterns (e.g.: `not' doesn't necessarily indicate negative sentiment), (iv) Manufacturing Industry: The test set has a rare chance of diverging from IID, and (v) Microbotics: Model computation and memory usage are important constraints. This is intended to allow users to map the metrics discussed and rankings observed to specific real-world applications that they have worked on, where top-ranked leaderboard models were found to fail during testing, if directly applied. Users are then asked to answer how the time, computation, and resources spent in using conventional leaderboards compares to using a customized, visually augmented leaderboard with the new metric(s) of their choice, in terms of: (i) selection of multiple models, (ii) custom-building of a new model from scratch, (iii) editing of an existing model, (iv) developing a testing strategy, and (v) creation, collection, and annotation of data. 

\begin{figure*}[t]
    \centering
    \includegraphics[width=\textwidth]{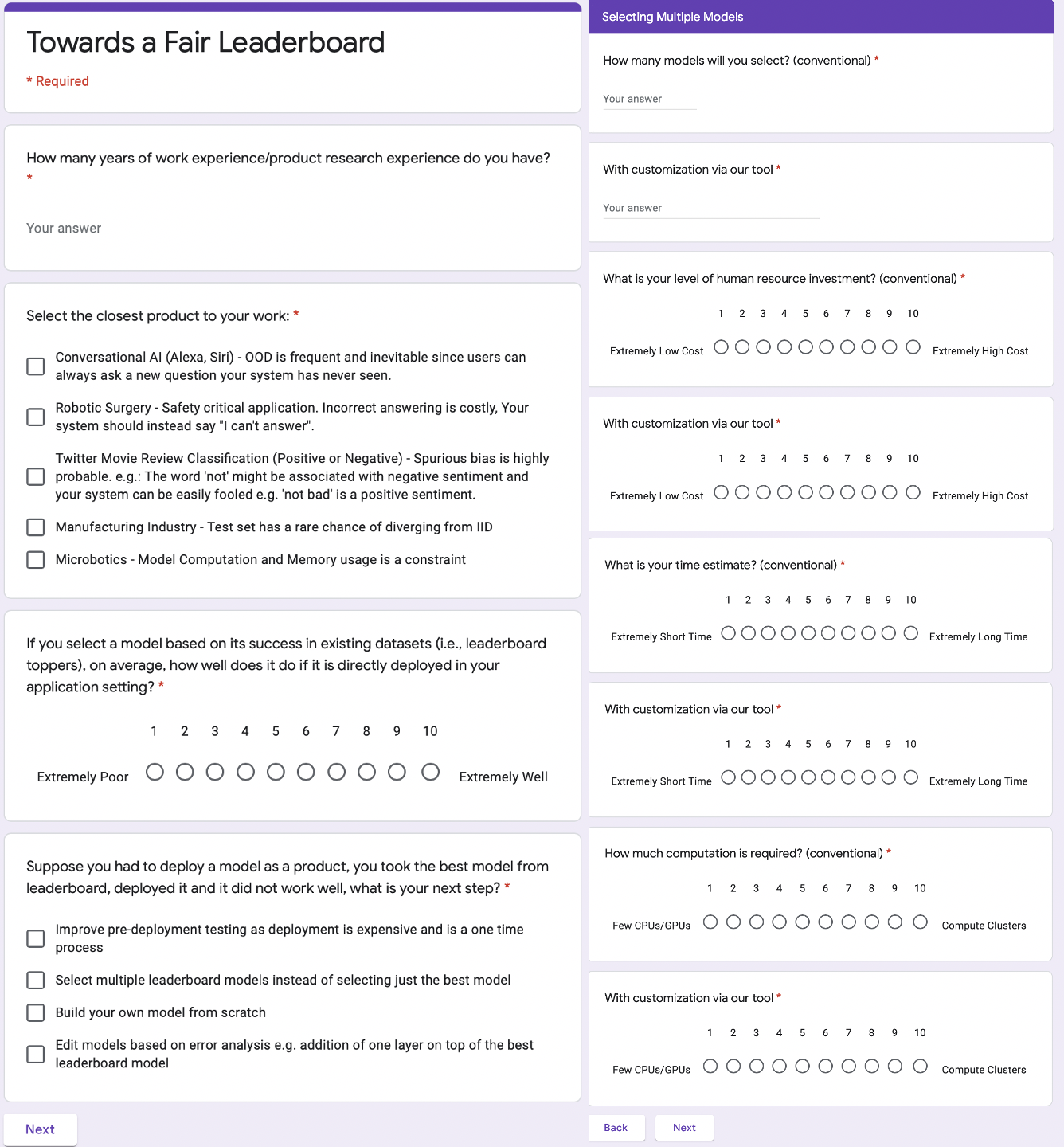}
    \caption{Google Form used for Survey}
    \label{fig:my_label1}
\end{figure*}

\begin{figure*}[t]
    \centering
    \includegraphics[width=\textwidth]{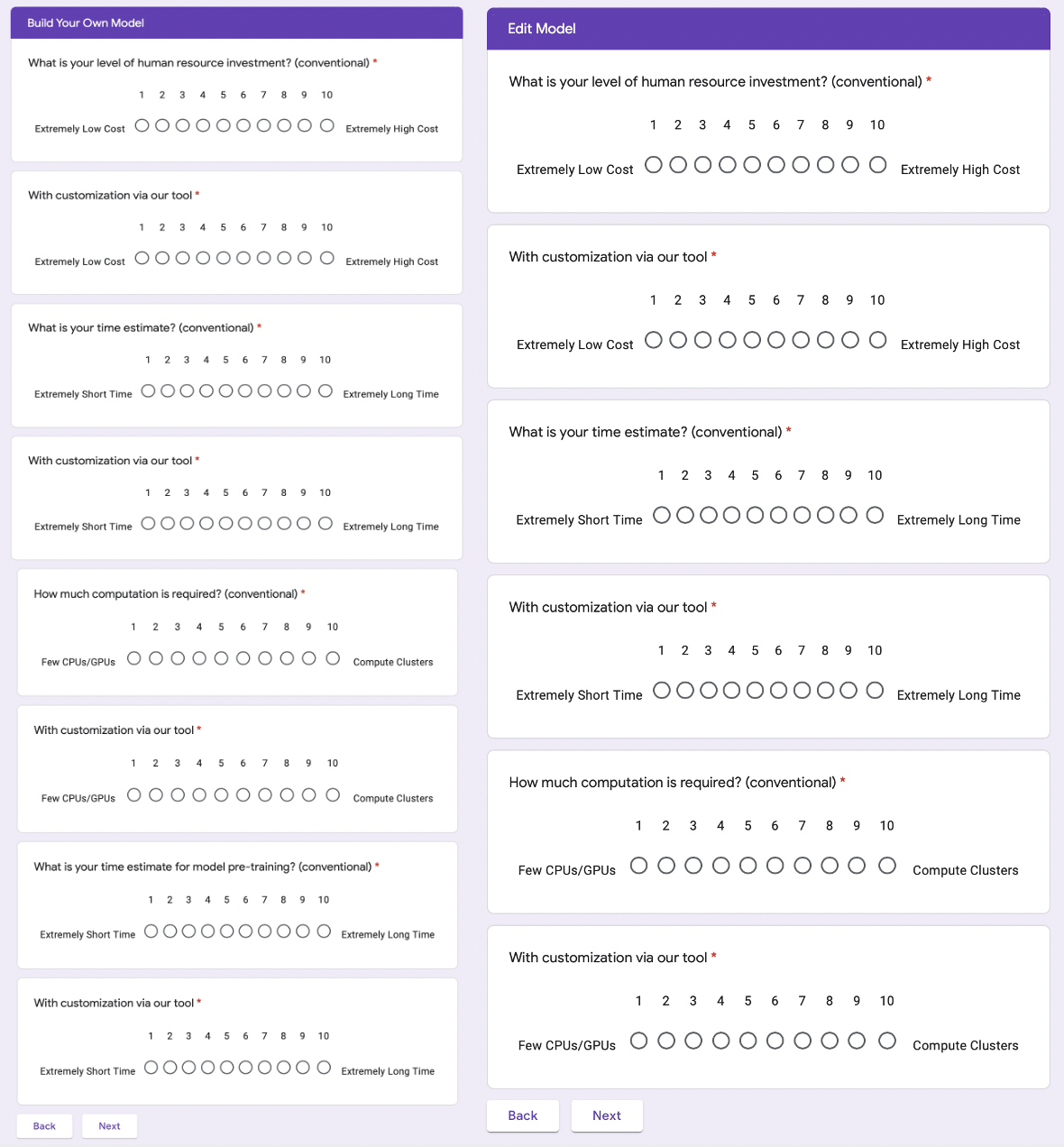}
    \caption{Google Form used for Survey}
    \label{fig:my_label2}
\end{figure*}

\begin{figure*}[t]
    \centering
    \includegraphics[width=\textwidth]{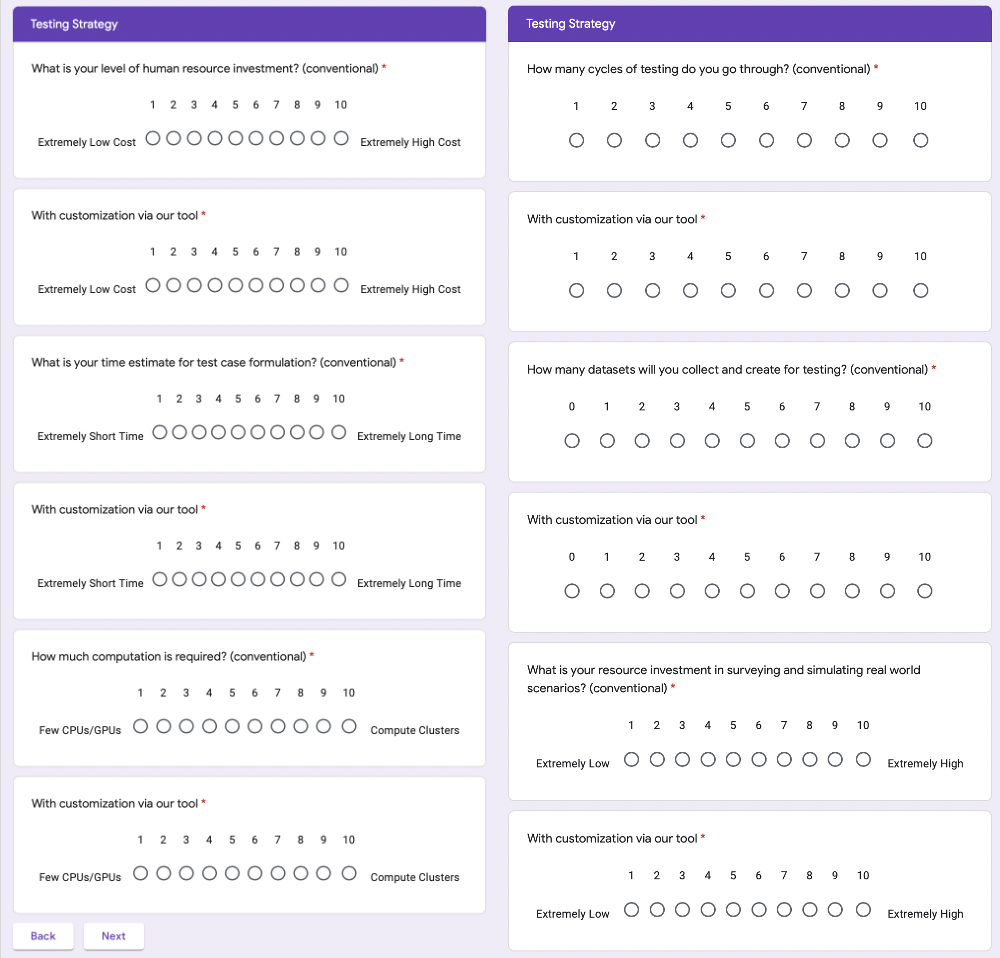}
    \caption{Google Form used for Survey}
    \label{fig:my_label3}
\end{figure*}

\begin{figure*}[t]
    \centering
    \includegraphics[width=\textwidth]{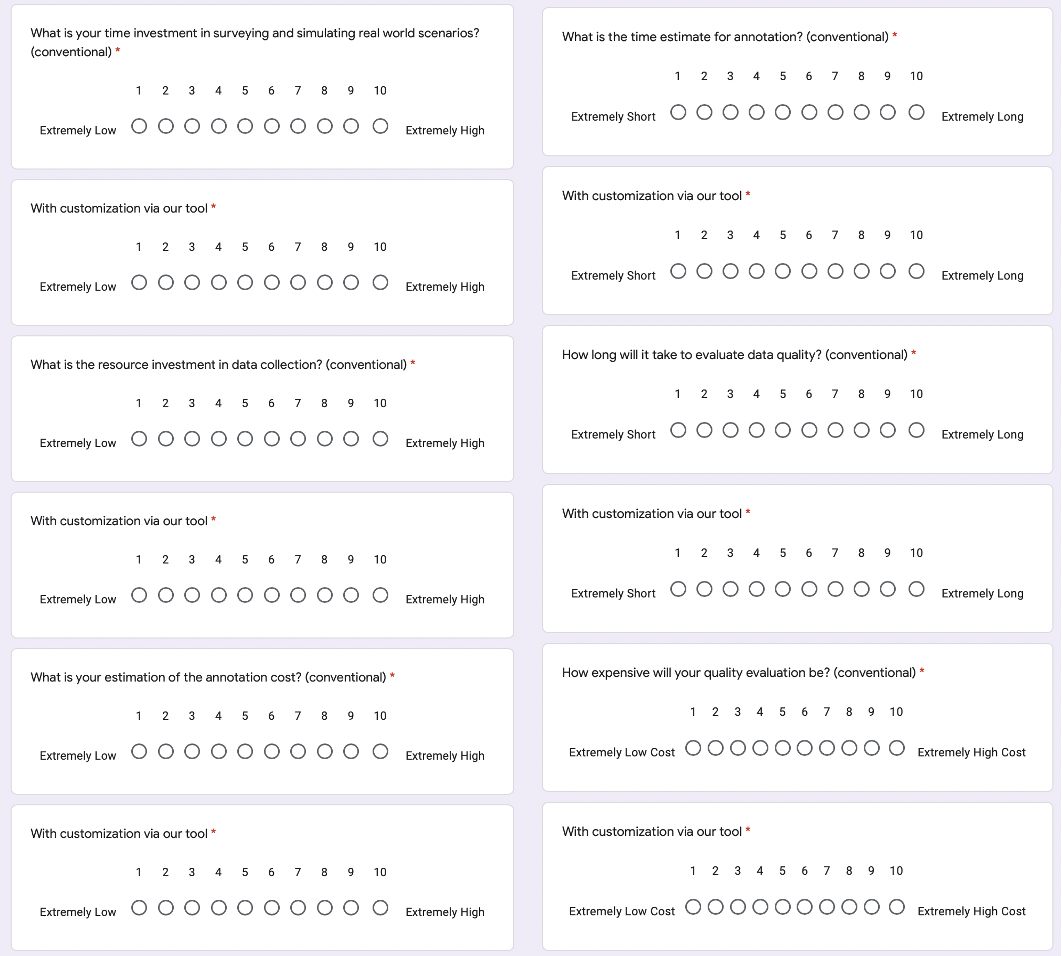}
    \caption{Google Form used for Survey}
    \label{fig:my_label4}
\end{figure*}

\begin{figure*}[t]
    \centering
    \includegraphics[width=0.7\textwidth,height=22cm]{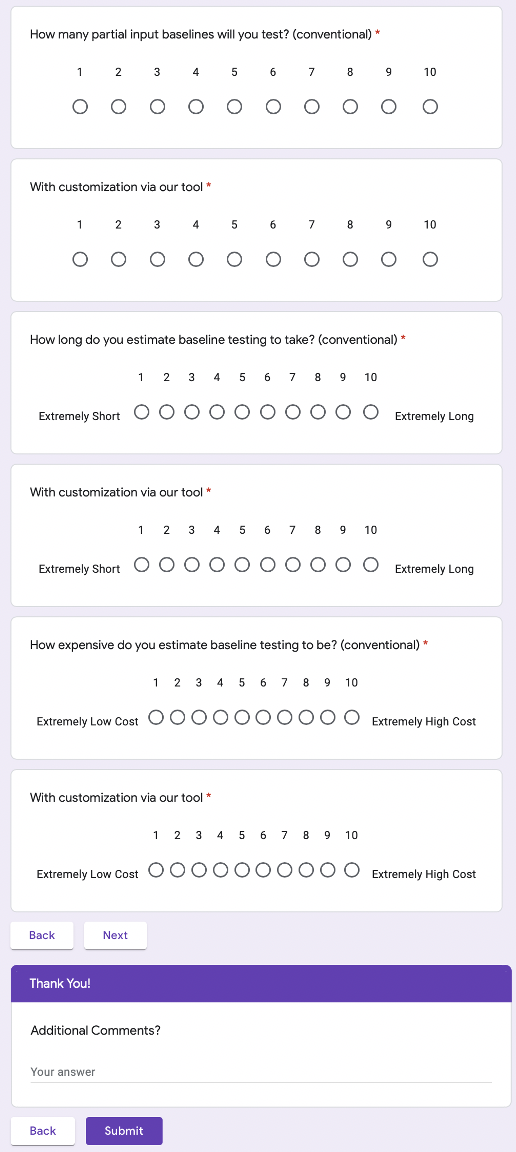}
    \caption{Google Form used for Survey}
    \label{fig:my_label5}
\end{figure*}

Please refer to Figures \ref{fig:my_label1},\ref{fig:my_label2},\ref{fig:my_label3},\ref{fig:my_label4},\ref{fig:my_label5}.


\end{document}